\documentclass[journal]{IEEEtran}
\usepackage{setspace}
\usepackage{amsmath}
\usepackage{amsfonts}
\usepackage{cjk}
\usepackage{indentfirst}
\usepackage{cite}
\usepackage{bm}
\usepackage{graphicx}
\usepackage{subfigure}
\usepackage{booktabs}
\usepackage{multirow}

\begin{document}

\title{Matrix Completion via Nonconvex Regularization: Convergence of the Proximal Gradient Algorithm}

\author{Fei Wen, 
        Rendong Ying,
        Peilin Liu, \textit{Senior Member}, \textit{IEEE},
        and Trieu-Kien Truong, \textit{Life Fellow}, \textit{IEEE}

\thanks{F. Wen, R. Ying, P. Liu and T.-K. Truong are with the Department of Electronic Engineering, Shanghai Jiao Tong University, Shanghai 200240, China (e-mail: wenfei@sjtu.edu.cn; rdying@sjtu.edu.cn; liupeilin@sjtu.edu.cn; truong@isu.edu.tw).}
}

\markboth{}
{Shell \MakeLowercase{\textit{et al.}}: Bare Demo of IEEEtran.cls for Journals}

\maketitle

\begin{abstract}
Matrix completion has attracted much interest in the past decade in machine learning and computer vision. For low-rank promotion in matrix completion, the nuclear norm penalty is convenient due to its convexity but has a bias problem. Recently, various algorithms using nonconvex penalties have been proposed, among which the proximal gradient descent (PGD) algorithm is one of the most efficient and effective. For the nonconvex PGD algorithm, whether it converges to a local minimizer and its convergence rate are still unclear. This work provides a nontrivial analysis on the PGD algorithm in the nonconvex case. Besides the convergence to a stationary point for a generalized nonconvex penalty, we provide more deep analysis on a popular and important class of nonconvex penalties which have discontinuous thresholding functions. For such penalties, we establish the finite rank convergence, convergence to restricted strictly local minimizer and eventually linear convergence rate of the PGD algorithm. Meanwhile, convergence to a local minimizer has been proved for the hard-thresholding penalty. Our result is the first shows that, nonconvex regularized matrix completion only has restricted strictly local minimizers, and the PGD algorithm can converge to such minimizers with eventually linear rate under certain conditions. Illustration of the PGD algorithm via experiments has also been provided. Code is available at https://github.com/FWen/nmc.
\end{abstract}


\IEEEpeerreviewmaketitle {}

\section{Introduction}
\label{sec:intro}

Matrix completion deals with the problem of recovering of a matrix from its partially observed (may be noisy) entries, which has attracted considerable interest recently [1]--[4]. The matrix completion problem arises in many applications in signal processing, image/video processing, and machine learning, such as rating value estimation in recommendation system [7], friendship prediction in social network, collaborative filtering [8], image processing [6], [10], video denoising [12], [13], system identification [14], multiclass learning [15], [16], and dimensionality reduction [17]. Specifically, the goal of matrix completion is to recover a matrix ${\bf{M}} \in {\mathbb{R}^{m \times n}}$ from its partially observed (incomplete) entries
\begin{equation} \label{I-1}
{{\bf{Y}}_{i,j}} = {{\bf{M}}_{i,j}},~~~~(i,j) \in \Omega
\end{equation}
where $\Omega \subset [1, \cdots ,m] \times [1, \cdots ,n]$ is a random subset.
Obviously, the completion of an arbitrary matrix is an ill-posed problem.
To make the problem well-posed, a commonly used assumption is that the underlying
matrix ${\bf{M}}$ comes from a restricted class, e.g., low-rank.
Exploiting the low-rank structure of the matrix is a powerful method.

Modeling the matrix completion problem as a low-rank matrix recovery problem,
a natural formulation is to minimize the rank of ${\bf{M}}$
under the linear constraint (\ref{I-1}) as
\begin{equation}\label{I-2}
\begin{split}
&\mathop {{\rm{minimize}}}\limits_{\bf{X}} {\rm{rank}}({\bf{X}})\\
&\textrm{subject to  } ~~{{ \mathcal{P}}_\Omega }({\bf{X}}) = {{\bf{Y}}_\Omega }
\end{split}
\end{equation}
where ${{\mathcal{ P}}_\Omega }:{\mathbb{R}^{m \times n}} \to {\mathbb{R}^{m \times n}}$
denotes projection onto the set $\Omega $, and ${{\bf{Y}}_\Omega } = {{\mathcal{ P}}_\Omega }({\bf{Y}})$.
While the nonconvex rank minimization problem (\ref{I-2})
is highly nonconvex and difficult to solve,
a popular convex relaxation method is to replace
the rank function by its convex envelope,
the nuclear norm ${\left\|  \cdot  \right\|_*}$,
\begin{equation}\label{I-3}
\begin{split}
&\mathop {{\rm{minimize}}}\limits_{\bf{X}} {\left\| {\bf{X}} \right\|_*}\\
&\textrm{subject to  } ~~{{ \mathcal{P}}_\Omega }({\bf{X}}) = {{\bf{Y}}_\Omega }.
\end{split}
\end{equation}
In most realistic applications, entry-wise noise is inevitable.
Taking entry-wise noise into consideration,
a robust variant of (\ref{I-3}) is
\begin{equation}\label{I-4}
\begin{split}
&\mathop {{\rm{minimize}}}\limits_{\bf{X}} {\left\| {\bf{X}} \right\|_*}\\
&\textrm{subject to  } ~~\left\| {{{\bf{Y}}_\Omega } - {{\mathcal{P}}_\Omega }({\bf{X}})} \right\|_F^2 \le \varepsilon
\end{split}
\end{equation}
where $\varepsilon > 0$ is the noise tolerance.
This constrained formulation (\ref{I-4})
can be converted into an unconstrained form as
\begin{equation}\label{I-5}
\mathop {{\rm{minimize}}}\limits_{\bf{X}} \frac{1}{2}\left\| {{{\bf{Y}}_\Omega } - {{\mathcal{P}}_\Omega }({\bf{X}})} \right\|_F^2 + \lambda {\left\| {\bf{X}} \right\|_*}
\end{equation}
where $\lambda > 0$ is a regularization parameter related to
the noise tolerance parameter $\varepsilon$ in (\ref{I-4}).
The unconstrained formulation is favorable in some applications
as existing efficient first-order convex algorithms,
such as alternative direction method of multipliers (ADMM) or
proximal gradient descent (PGD) algorithm, can be directly applied.
Even in the noise free case, the solution of (\ref{I-5}) can accurately approach
that of (\ref{I-3}) via choosing a sufficiently small value of $\lambda $,
since the solution of (\ref{I-5}) satisfies ${\left\| {{{\bf{Y}}_\Omega } - {{\mathcal{P}}_\Omega }({\bf{X}}))} \right\|_F} \to 0$
as $\lambda  \to {\rm{0}}$. The problems (\ref{I-3}) and (\ref{I-4}) can be recast
into semi-definite program (SDP) problems and solved to global minimizer
by well-established SDP solvers when the matrix dimension is not large.
For problems with larger size, more efficient first-order algorithms
have been developed based on the formulation (\ref{I-5}), e.g.,
variants of the proximal gradient method [19], [20].

Besides the tractability of the convex formulations
(\ref{I-3})--(\ref{I-5}) employing nuclear norm,
theoretical guarantee provided in [1], [2], [21], [22] demonstrated that under certain conditions, e.g.,
when the low-rank matrix ${\bf{M}}$ satisfies an incoherence condition
and the observed entries are uniformly randomly sampled, ${\bf{M}}$ can
be exactly recovered from a small portion of its entries with high
probability by using the nuclear norm regularization. However,
the nuclear norm regularization has a bias problem and would introduce
bias to the recovered singular values [23]--[25]. To alleviate the bias
problem and achieve better recovery performance, a nonconvex low-rank penalty,
such as the Schatten-$q$ norm (which is in fact the $\ell_q$ norm of the matrix
singular values with $0 < q < 1$), smoothly clipped absolute deviation (SCAD),
minimax concave (MC), or firm-thresholding penalty can be used.
In the past a few years, nonconvex regularization
has shown better performance over convex regularization
in many sparse and low-rank recovery involved applications.
These applications include compressive sensing,
sparse regression, sparse demixing, sparse covariance and precision matrix estimation,
and robust principal component analysis [9], [26].

In this work, we consider the following formulation for matrix completion
\begin{equation}\label{I-6}
\mathop {{\rm{minimize}}}\limits_{\bf{X}} F({\bf{X}}): = \frac{1}{2}\left\| {{{\bf{Y}}_\Omega } - {{\mathcal{P}}_\Omega }({\bf{X}})} \right\|_F^2 + \lambda \bar R({\bf{X}})
\end{equation}
where $\bar R$ is a generalized nonconex low-rank promotion penalty.
For the particular case of $\bar R$ being the nuclear norm, i.e.,
$\bar R( \cdot ) = {\left\|  \cdot  \right\|_*}$, this formulation reduces to (\ref{I-5}).
Existing works considering the nonconvex formulation (\ref{I-6}) include [27]--[31].
In [27], [28], the Schatten-$q$ norm has been considered and PGD methods have been proposed.
In [29], using a smoothed Schatten-$q$ norm, an iteratively reweighted algorithm has been
designed for (\ref{I-6}), which involves solving a sequence of linear equations.
Another iteratively reweighted algorithm for Schatten-$q$ norm regularized matrix
minimization problem with a generalized smooth loss function has been investigated in [30].
More recently in [31], $\bar R$ being the MC penalty has been considered and an ADMM algorithm has been developed.

Besides, for the linearly constrained formulation, an iterative algorithm employing Schatten-$q$ norm,
which monotonically decreasing the objective, has been proposed in [32]. Meanwhile, a truncated nuclear
norm has been used in [33]. Then, robust matrix completion using Schatten-$q$ regularization has been considered in [34].
Moreover, it has been shown in [35] that, the sufficient condition for reliable recovery of
Schatten-$q$ norm regularization is weaker than that of nuclear norm regularization.

Among the nonconvex algorithms for the problem (\ref{I-6}),
only subsequence convergence of the methods [27]--[31] have been proved.
In fact, based on the recent convergence results for nonconvex and nonsmooth optimization [36]--[38],
global convergence of the PGD algorithm [27], [28] and the ADMM algorithm [31]
to a stationary point can be guaranteed under some mild conditions.
However, for a nonconvex $\bar R$, whether these algorithms converge to a local minimizer is still unclear.
Meanwhile, for the problem (\ref{I-6}), linear convergence rate of the PGD algorithm has been established
when $\bar R$ is the nuclear norm under certain conditions [39], [40],
but the convergence rate of PGD in the case of a nonconvex $\bar R$ is still an open problem.

To address these problems, this work provides a thorough
analysis on the PGD algorithm for the matrix completion problem (\ref{I-6})
using a generalized nonconvex penalty. The main contributions are as follows.

\subsection{Contribution}

First, we derived some properties on the gradient and
Hessian of a generalized low-rank penalty,
which are important for the convergence analysis.
Then, for a popular and important class of nonconvex penalties
which have discontinuous thresholding functions,
we have established the following
convergence properties for the PGD algorithm under certain conditions:

\hangindent 2.2em
{1)} rank convergence within finitely many iterations;

\hangindent 2.2em
{2)} convergence to a restricted strictly local minimizer;

\hangindent 2.2em
{3)} convergence to a local minimizer for the hard-thresholding penalty;

\hangindent 2.2em
{4)} an eventually linear convergence rate.

\noindent
As the singular value thresholding function is
implicitly dependent on the low-rank matrix,
the derivation is nontrivial. Finally, illustration
of the PGD algorithm via inpainting experiments has been provided.

It is worth noting that, there exist a line of recent
works on factorization based nonconvex algorithms, e.g., [5], [11], [18].
It has been shown that the nonconvex objective function has no spurious local minimum,
and efficient nonconvex optimization algorithms can converge to local minimum.
While these works focus on matrix factorization based methods,
this work considers the general matrix completion problem (\ref{I-6}).
Our result is the first explains that the nonconvex
matrix completion problem (\ref{I-6}) only have restricted
strictly local minimum, and the PGD algorithm can converge
to such minimum with eventually linear rate under certain conditions.

\textit{Outline:} The rest of this paper is organized as follows.
Section II introduces the proximity operator for generalized nonconvex penalty,
and reviews the PGD algorithm for matrix completion.
Section III provides convergence analysis of the PGD algorithm.
Section IV provides experimental results on inpainting.
Finally, section V ends the paper with concluding remarks.

\renewcommand\arraystretch{1.5}
\begin{table*}[!t]
\caption{Proximity operator for some popular regularization penalties.}
\footnotesize
\centering
\newcommand{\tabincell}[2]{\begin{tabular}{@{}#1@{}}#2\end{tabular}}
\begin{tabular}{|p{3.5cm}|p{6cm}|p{7cm}|}
\hline
Penalty name & Penalty formulation & Proximity operator \\

\hline
(i) Hard thresholding & ${R }(x) =  {\rm{|}}x{{\rm{|}}_0}$ & ${P_{R,\eta }}(t) = \left\{ {\begin{array}{*{20}{l}}
{0,}&{|t| \leq \sqrt {2/\eta } }\\
{t,}&{|t| \geq \sqrt {2/\eta } }
\end{array}} \right.$ \\

\hline
(ii)  Soft thresholding & ${R }(x) = {\rm{|}}x{\rm{|}}$ & ${P_{R,\eta }}(t) = {\rm{sign}}(t)\max \left\{ {|t| - 1/\eta ,0} \right\}$ \\

\hline
(iii) $\ell_q$-norm  & ${R }(x) = {\rm{|}}x{{\rm{|}}^q}$, $0 < q < 1$  &
\tabincell{c}{${P_{R,\eta }}(t) = \left\{ {\begin{array}{*{20}{l}}
{0,} & {{\rm{|}}t{\rm{|}} \leq \tau }\\
{{{\rm{sign}}(t){h^{ - 1}}({\rm{|}}t{\rm{|}})},} & {{\rm{|}}t{\rm{|}} \geq \tau }
\end{array}} \right.$\\
where $h(x) = q{x^{q - 1}}/\eta  + x$, $\tau  = {\beta _\eta } + q\beta _\eta ^{q - 1}/\eta$,\\
 ${\beta _\eta } = {[2(1 - q)/\eta ]^{1/(2 - q)}}$~~~~~~~~~~~~~~~~~~~~~~~~~~~}\\
\hline
\end{tabular}
\end{table*}
\renewcommand\arraystretch{1}

\textit{Notations}: For a matrix ${\bf{X}}\in {\mathbb{R}^{m \times n}}$, ${\rm{rank}}({\bf{X}})$, ${\rm{tr}}({\bf{X}})$,
${\left\| {\bf{X}} \right\|_F}$ and $\mathcal{R}({\bf{X}})$ stand for the rank, trace,
Frobenius norm and range space of ${\bf{X}}$, respectively,
whilst ${\sigma _i}({\bf{X}})$ denotes the $i$-th largest singular value, and
\begin{align}\notag
{\boldsymbol{\sigma }}({\bf{X}})&:= {[{\sigma _1}({\bf{X}}), \cdots ,{\sigma _{\min (m,n)}}({\bf{X}})]^T}\\ \notag
{{\boldsymbol{\sigma }}_r}({\bf{X}})&:= {[{\sigma _1}({\bf{X}}), \cdots ,{\sigma _r}({\bf{X}})]^T}\\ \notag
{{\boldsymbol{\sigma }}_{r \bot }}({\bf{X}})&:= {[{\sigma _{r + 1}}({\bf{X}}), \cdots ,{\sigma _{\min (m,n)}}({\bf{X}})]^T}.
\end{align}
For a symmetric real matrix ${\bf{X}}$,
${\lambda _{\max }}({\bf{X}})$ and ${\lambda _{\min }}({\bf{X}})$ respectively denote the maximal and minimal eigenvalues,
whilst ${\boldsymbol{\lambda }}({\bf{X}})$ contains the descendingly ordered eigenvalues.
${\bf{X}}\succeq{\bf{0}}$ and ${\bf{X}} \succ {\bf{0}}$ mean that ${\bf{X}}$ is semi-definite and positive definite, respectively.
${\bf{X}}(i,j)$ denotes the $(i,j)$-th element. ${\rm{vec}}( \cdot )$ is the ``vectorization'' operator stacking
the columns of the matrix one below another.
${\bf{diag}}({\bf{v}})$ represents the diagonal matrix generated by the vector ${\bf{v}}$,
${\rm{diag}}({\bf{X}})$ represents the vector containing the diagonal elements of ${\bf{X}}$.
${\left\|  \cdot  \right\|_2}$ denotes the Euclidean norm.
$\odot $ and $\otimes$ denote the Hadamard and Kronecker product, respectively.
$\langle\cdot,\cdot\rangle$ and ${( \cdot )^T}$ denote the inner product and transpose, respectively.
${\rm{sign}}( \cdot )$ denotes the sign of a quantity with ${\rm{sign}}(0){\rm{ = }}0$.
${{\bf{I}}_m}$ is an $m \times m$ identity matrix.
${\bf{0}}$ is a zero vector or matrix with a proper size.

\section{Proximity Operator and Proximal Gradient Algorithm}

This section introduces the proximity operator for nonconvex
regularization and the PGD algorithm for the matrix completion
problem (\ref{I-6}).

\subsection{Proximal Operator for Nonconvex Penalties}

For a proper and lower semicontinuous penalty function $R$,
the corresponding proximity operator is defined as
\begin{equation}\label{II-7}
{P_{R,\eta }}(t) = \arg \mathop {\min }\limits_x \left\{ {R(t) + \frac{\eta }{2}{{(x - t)}^{\rm{2}}}} \right\}
\end{equation}
where $\eta > 0$ is a penalty parameter.

Table I shows several popular penalties along with their thresholding functions.
The proximal minimization problem (\ref{II-7}) for many popular nonconvex penalties can be
computed in an efficient manner. The hard-thresholding is a natural selection for sparsity promotion,
while the soft-thresholding is of the most popular due to its convexity.
The $\ell_q$ penalty with $0<q<1$ bridges the gap between the hard- and soft-thresholding penalties.
Except for two known cases of $q = \frac{1}{2}$ and $q = \frac{2}{3}$,
the proximity operator of the ${\ell _q}$ penalty does not have a closed-form expression,
but it can be efficiently computed by an iterative method.
Moreover, there also exist other nonconvex penalties,
including the $q$-shrinkage [41]--[42], SCAD [43], MC [44] and
firm thresholding [45].

As shown in Fig. 1, the soft-thresholding imposes a
constant shrinkage on the parameter when the parameter magnitude
exceeds the threshold, and, thus, has a bias problem.
The hard- and SCAD thresholding are unbiased for large parameter.
The other nonconvex thresholding functions are sandwiched between
the hard- and the soft-thresholding,
which can mitigate the bias problem of the soft-thresholding.
For a generalized nonconvex penalty, we make the following assumptions.

\begin{figure}[!t]
 \centering
 \includegraphics[scale = 0.35]{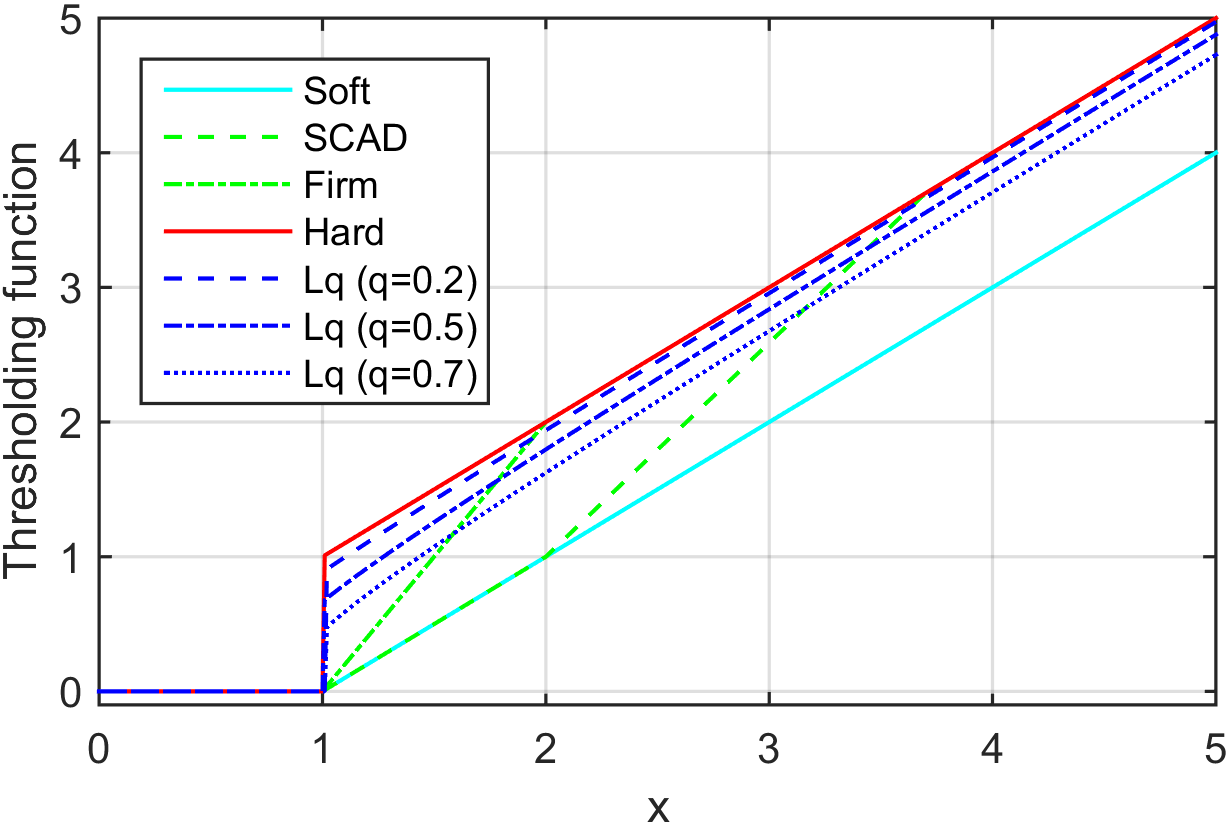}
\caption{Thresholding/shrinkage function output (with a same threshold).}
 \label{figure1}
\end{figure}

\textit{Assumption 1:} ${R}$ is an even folded concave function,
which satisfies the following conditions:

\hangindent 2.2em
(i) ${R}$ is non-decreasing on $[0,\infty )$ with ${R}(0) = 0$;

\hangindent 2.2em
(ii) for any $t > 0$, there exists a $c > 0$ such that ${R}(|x|) \ge c{x^2}$ for any $|x| \in [0,t]$;

\hangindent 2.2em
(iii) ${R}$ is ${C^2}$ on $( - \infty ,0) \cup (0,\infty )$, and ${R''}\leq 0$ on $ (0,\infty )$;

\hangindent 2.2em
(iv) the first-order derivative ${R'}$ is convex on $(0,\infty )$ and $\mathop {\lim }\limits_{|x| \to \infty } {R'}(|x|)/|x| = 0$.

This assumption implies that $R$ is coercive,
weakly sequential lower semi-continuous in $\ell^2$,
and responsible for sparsity promotion.

\subsection{Generalized Singular Value Thresholding}

For a matrix ${\bf{X}} \in {\mathbb{R}^{m \times n}}$,
low-rank inducing on ${\bf{X}}$ can be achieved via
sparsity inducing on the singular values as
\begin{equation}\label{II-8}
\bar R({\bf{X}}): = R({\boldsymbol{\sigma }}({\bf{X}})) = \sum\nolimits_{i = 1}^{} {R({\sigma _i}({\bf{X}}))}
\end{equation}
where $R$ is a sparsity inducing penalty.
For the particular cases of $R$ being the ${\ell _0}$, ${\ell _q}$ and ${\ell _1}$ norm,
$\bar R({\bf{X}})$  become the rank, Schatten-$q$ norm and nuclear norm of ${\bf{X}}$, respectively.
For such a low-rank penalty, define the corresponding proximal operator
\begin{equation}\label{II-9}
{\bar P_{\bar R,\eta }}({\bf{T}}) = \arg \mathop {\min }\limits_{\bf{X}} \left\{ {\bar R({\bf{X}}) + \frac{\eta }{2}\left\| {{\bf{X}} - {\bf{T}}} \right\|_F^2} \right\}.
\end{equation}

\textit{Property 1. [Generalized singular value thresholding]:}
Let ${\bf{T}} = {\bf{Udiag}}({\boldsymbol{\sigma }}({\bf{T}})){{\bf{V}}^T}$
be any full singular value decomposition (SVD) of ${\bf{T}}$,
where ${\bf{U}} \in {\mathbb{R}^{m \times m}}$ and ${\bf{V}} \in {\mathbb{R}^{n \times n}}$
contain the left and right singular vectors, respectively.
Then, the proximal minimization problem (\ref{II-9}) is solved by the
singular-value thresholding operator
\begin{equation}\label{II-10}
{\bar P_{\bar R,\eta }}({\bf{T}}) = {\bf{U}}  {\bf{diag}}\left\{ {{P_{R,\eta }}({\boldsymbol{\sigma }}({\bf{T}}))} \right\}  {{\bf{V}}^T}
\end{equation}
where
\[{P_{R,\eta }}({\boldsymbol{\sigma }}({\bf{T}})) = {[{P_{R,\eta }}({\sigma _1}({\bf{T}})), \cdots ,{P_{R,\eta }}({\sigma _{\min (m,n)}}({\bf{T}}))]^T}.\]

Although this property can be derived via straightforwardly extending Lemma 1 in [7],
we provide here a completely different but more intuitive derivation of it.
Assume that the minimizer ${{\bf{X}}^*}$ of (\ref{II-9}) is of rank $r$
with any truncated SVD ${{\bf{X}}^*} = {{\bf{U}}^*}{{\bf{\Sigma }}^*}{{\bf{V}}^{*T}}$,
where ${{\bf{\Sigma }}^*}{{ = }}{\bf{diag}}({{\boldsymbol{\sigma }}_r}({{\bf{X}}^*}))$.
Then, the objective in (\ref{II-9}) can be equivalently rewritten as
\begin{equation}\label{II-11}
T({\bf{X}}) := R({{\boldsymbol{\sigma }}_r}({\bf{X}})) + \frac{\eta }{2}\left\| {{\bf{X}} - {\bf{T}}} \right\|_F^2.
\end{equation}
By Assumption 1, $R$ is differential on $(0, + \infty )$,
hence, $T$ is differential with respective to rank-$r$ matrix ${\bf{X}}$.
Denote
\[{{\bf{\Sigma }}^*}^\prime  = {\bf{diag}}({[R'({\sigma _1}({{\bf{X}}^*})), \cdots ,R'({\sigma _r}({{\bf{X}}^*}))]^T})\]
where $R'$ is the first-order derivative of $R$, we have (see Appendix A)
\begin{equation}\label{II-12}
{\nabla _{\bf{X}}}T({{\bf{X}}^*}) = {{\bf{U}}^*}{{\bf{\Sigma }}^*}^\prime {{\bf{V}}^{*T}} + \eta ({\bf{X^*}} - {{\bf{U}}^*}{{\bf{U}}^{*T}}{\bf{TV}}{{\bf{V}}^{*T}}).
\end{equation}
Let ${\nabla _{\bf{X}}}T({{\bf{X}}^*}) = {\bf{0}}$,
and use ${{\bf{U}}^{*T}}{{\bf{U}}^*} = {{\bf{I}}_r}$,
${{\bf{V}}^{*T}}{{\bf{V}}^*} = {{\bf{I}}_r}$,
it follows from (\ref{II-12}) that
\[{{\bf{\Sigma }}^*}^\prime  + \eta {{\bf{\Sigma }}^*} - \eta {{\bf{U}}^{*T}}{\bf{T}}{{\bf{V}}^*} = {\bf{0}}.\]
Since ${{\bf{\Sigma }}^*}^\prime $ and ${{\bf{\Sigma }}^*}$ are diagonal,
and the columns of ${{\bf{U}}^*}$ (also ${{\bf{V}}^*}$) are orthogonal,
it is easy to see that there exists a full SVD ${\bf{T}} = {\bf{U\Sigma }}{{\bf{V}}^T}$ such that
\begin{equation}\label{II-13}
{\bf{U}} = [{{\bf{U}}^*},{\bf{U}}_ \bot ^*]~~~~ {\rm{and}}~~~~ {\bf{V}} = [{{\bf{V}}^*},{\bf{V}}_ \bot ^*].
\end{equation}
Substituting these relations into (\ref{II-11}) yields
\begin{equation}\label{II-14}
T({\bf{X}}^*) = R({{\boldsymbol{\sigma }}_r}({{\bf{X}}^*})) + \frac{\eta }{2}\left\| {{{\boldsymbol{\sigma }}_r}({{\bf{X}}^*}) - {{\tilde{\boldsymbol{ \sigma }}}_{T,r}}} \right\|_2^2
\end{equation}
where ${\tilde{\boldsymbol{ \sigma }}_{T,r}}$ contains $r$ singular values of ${\bf{T}}$.
As (\ref{II-14}) is separable, ${\{ {\sigma _i}({{\bf{X}}^*})\} _{1 \le i \le r}}$
can be solved element-wise as (\ref{II-7}), i.e.,
${\sigma _i}({{\bf{X}}^*}) = {P_{R,\eta }}({\tilde{\boldsymbol{ \sigma }}_{T,r}}(i))$.
Further, $R$ is nondecreasing on $(0, + \infty )$ by Assumption 1,
hence ${P_{R,\eta }}(x) \le {P_{R,\eta }}(y)$ for any $0 < x \le y$.
Thus, ${\tilde{\boldsymbol{ \sigma }}_{T,r}}$ must contain the $r$ largest singular
values of ${\bf{T}}$ with a same descending order as ${{\boldsymbol{\sigma }}_r}({{\bf{X}}^*})$,
i.e., ${\tilde{\boldsymbol{ \sigma }}_{T,r}} = {{\boldsymbol{ \sigma }}_r}({\bf{T}}) = {[{\sigma _1}({\bf{T}}), \cdots ,{\sigma _r}({\bf{T}})]^T}$.
Consequently, we have ${{\boldsymbol{\sigma }}_r}({{\bf{X}}^*}) = {P_{R,\eta }}({{\boldsymbol{\sigma }}_r}({\bf{T}}))$,
which together with ${P_{R,\eta }}({[{\sigma _{r + 1}}({\bf{T}}), \cdots ,{\sigma _{\min (m,n)}}({\bf{T}})]^T}) = {\bf{0}}$
and (\ref{II-13}) results in (\ref{II-10}).

\subsection{PGD Algorithm for Matrix Completion}

PGD is a powerful optimization algorithm suitable for many large-scale problems arising in signal/image processing, statistics and machine learning. It can be viewed as a variant of majorization minimization algorithms which has a special choice for the quadratic majorization. Let
\[G({\bf{X}}): = \frac{1}{2}\left\| {{{\bf{Y}}_\Omega } - {{\mathcal{P}}_\Omega }({\bf{X}}))} \right\|_F^2.\]
The core idea of the PGD algorithm is to consider a linear
approximation of $G$ at the $(k+1)$-th iteration at a given point ${{\bf{X}}^k}$ as
\begin{equation}\label{II-15}
\begin{split}
{F_L}({\bf{X}};{{\bf{X}}^k})= G({{\bf{X}}^k}) &+ \left\langle {{\bf{X}} - {{\bf{X}}^k},\nabla G({{\bf{X}}^k})} \right\rangle \\
 &+ \frac{L}{2}\left\| {{\bf{X}} - {{\bf{X}}^k}} \right\|_F^2 + \lambda \bar R({\bf{X}})
\end{split}
\end{equation}
where $\nabla G({{\bf{X}}^k}) = {{\mathcal{P}}_\Omega }({{\bf{X}}^k}) - {{\bf{Y}}_\Omega }$
and $L > 0$ is a proximal parameter. Then, minimizing
${F_L}({\bf{X}};{{\bf{X}}^k})$ is a form of the proximity operator (\ref{II-9}) as
\begin{equation}\label{II-16}
{{\bf{X}}^{k + 1}} = {\bar P_{\bar R,L/\lambda }}\left( {{{\bf{X}}^k} - \frac{1}{L}\nabla G({{\bf{X}}^k})} \right)
\end{equation}
which can be computed as (\ref{II-10}).

In the PGD algorithm, the dominant computational load in each iteration is the SVD calculation.
To further improve the efficiency of the algorithm and make it scale well for large-scale problems,
the techniques such as approximate SVD or PROPACK [7], [19] can be adopted.

\section{Convergence Analysis}

This section investigates the convergence properties of the
PGD algorithm with special consideration on the class of
nonconvex penalties which have discontinuous thresholding functions.
First, we make some assumptions on the discontinuous property of such
threshoding functions.

\textit{Assumption 2:}
${R}$ satisfies Assumption 1,
and the corresponding proximity operator has a formulation as
\begin{equation}\label{III-17}
{P_{{R},\eta }}(t) = \left\{ {\begin{array}{*{20}{l}}
{0,}&{{\rm{|}}t{\rm{|}} \le {\tau _\eta }}\\
{{\rm{sign}}(t)\rho _\eta ^{ - 1}(|t|),}&{{\rm{|}}t{\rm{|}} \ge {\tau _\eta }}
\end{array}} \right.
\end{equation}
where ${\rho _\eta }$ is defined on $\mathbb{R}_+$ as
${\rho _\eta }:x \mapsto {R'}(x)/\eta  + x$, for any $\eta  > 0$ and $x > 0$.
${\tau _\eta } > 0$ is the threshold point given by ${\tau _\eta } = {\rho _\eta }({\beta _\eta })$.
${\beta _\eta } = \rho _\eta ^{ - 1}({\tau _\eta })>0$ is the ``jumping''
size at the threshold point. ${P_{R,\eta }}(t)$ is continuous
on $\{ |t| \ne {\tau _\eta }\} $ and the range of
${P_{{R},\eta }}(t)$ is $( - \infty , - {\beta _\eta }] \cup \{ 0\}  \cup [{\beta _\eta }, + \infty )$.

A significant property of such a nonconvex penalty is its jumping discontinuity.
Typical nonconvex penalties satisfying this discontinuous property include the
${\ell _0}$, ${\ell _q}$, and log-$q$ penalties.

In the analysis, the Kurdyka-Lojasiewicz (KL) property of the
objective function is used. In the convergence analysis, based on
a ``uniformization'' result [36], using the KL property
can considerably predigest the main arguments and avoid involved
induction reasoning.

\textit{Definition 1. [KL property]:}
For a proper function $f:{\mathbb{R}^n} \to \mathbb{R}$ and any ${x_0} \in {\rm{dom}}\partial f$,
if there exists $\eta  > 0$, a neighborhood $\mathcal{V}$ of ${x_0}$ and a
continuous concave function $\varphi :[0,\eta ) \to { \mathbb{R}_+ }$ such that:

\hangindent 2.2em
(i) $\varphi (0) = 0$ and $\varphi$ is continuously differentiable on $(0,\eta )$ with positive derivatives;

\hangindent 2.2em
(ii) for all $x \in \mathcal{V}$ satisfying $f({x_0}) < f(x) < f({x_0}) + \eta$, it holds that $\varphi '(f(x) - f({x_0})){\rm{dist}}(0,\partial f(x)) \ge 1$;

\noindent
then $f$ is said to have the KL property at $x_0$.
Further, if a proper closed function $f$ satisfies
the KL property at all points in ${\rm{dom}}\partial f$,
it is called a KL function.

Furthermore, we define the restricted strictly local minimizer as follows.
Let ${ \mathcal{P}}_\Omega ^ \bot :{\mathbb{R}^{m \times n}} \to {\mathbb{R}^{m \times n}}$
denote the projection onto the complementary set of $\Omega $.

\textit{Definition 2. [Restricted strictly local minimizer]:}
For a proper function $f:{\mathbb{R}^{m \times n}} \to {\mathbb{R}}$, any ${{\bf{X}}^*} \in {\rm{dom}}\partial f$
and a subset $\Omega  \subset [1, \cdots ,m] \times [1, \cdots ,n]$,
if there exists a neighborhood $\mathcal{V}$ of ${{\bf{X}}^*}$ such that for any ${\bf{X}} \in \mathcal{V}$,
\[f({\mathcal{P}_\Omega }({\bf{X}}) + \mathcal{P}_\Omega ^ \bot ({{\bf{X}}^*})) > f({{\bf{X}}^*})\]
${{\bf{X}}^*}$ is said to be a $\Omega $-restricted strictly local minimizer of $f$.

It is obvious that, if ${{\bf{X}}^*}$ is a strictly local minimizer of $f$,
then ${{\bf{X}}^*}$ is a $\Omega $-restricted strictly local minimizer of $f$, but not vice versa.

Meanwhile, we provide three  lemmas needed in later analysis.
The first lemma is on the distance between the singular values of two matrices.

\textit{{Lemma 1:}} For two matrices ${\bf{A}} \in {\mathbb{R}^{m \times n}}$ and ${\bf{B}} \in {\mathbb{R}^{m \times n}}$, it holds
\[\left\| {{\boldsymbol{\sigma }}({\bf{A}}) - {\boldsymbol{\sigma }}({\bf{B}})} \right\|_2^2 \le \left\| {{\bf{A}} - {\bf{B}}} \right\|_F^2.\]

This result can be directly derived by extending the Hoffman-Wielandt Theorem \cite{47},
which indicates that the ``distance'' between the respective
singular values of two matrices is bounded by the ``distance'' between the matrices.

The following two lemmas present some properties of the
gradient and Hessian of a generalized low-rank penalty [46]
(the derivation is also provided here in Appendices A and B).

\textit{{Lemma 2:}}
For a matrix ${\bf{X}} \in {\mathbb{R}^{m \times n}}$ of rank $r$,
$r \le \min (m,n)$, with any truncated SVD ${\bf{X}} = {\bf{U\Sigma V}}^T$,
${\bf{\Sigma }} = {\bf{diag}}({{\boldsymbol{\sigma }}_r}({\bf{X}}))$,
${\bf{U}} \in {\mathbb{R}^{m \times r}}$ and
${\bf{V}} \in {\mathbb{R}^{n \times r}}$ contains the corresponding singular vectors.
Suppose that $R$ is $C^2$ on $(0, + \infty )$, denote
\[{\bf{\Sigma'}} = {\bf{diag}}(R'({\sigma _1}({\bf{X}})), \cdots ,R'({\sigma _r}({\bf{X}})))\]
\[{\bf{\Sigma''}} = {\bf{diag}}(R''({\sigma _1}({\bf{X}})), \cdots ,R''({\sigma _r}({\bf{X}}))).\]
Then, ${\nabla _{\bf{X}}}R({{\boldsymbol{\sigma }}_r}({\bf{X}})) = {\bf{U\Sigma '}}{{\bf{V}}^T}$ and
\begin{equation}\notag
\begin{split}
\nabla _{\bf{X}}^2R({{\boldsymbol{\sigma }}_r}({\bf{X}}))= \frac{1}{2}{{\bf{K}}_{nm}}\big[ &({\bf{U\Sigma ''}}{{\bf{V}}^T}) \otimes ({\bf{V}}{{\bf{U}}^T})\\
& + ({\bf{U}}{{\bf{V}}^T}) \otimes ({\bf{V\Sigma ''}}{{\bf{U}}^T}) \big]
\end{split}
\end{equation}
where ${{\bf{K}}_{nm}}$ is a commutation matrix defined as
${\rm{vec}}({\bf{A}}) = {{\bf{K}}_{nm}}{\rm{vec}}({{\bf{A}}^T})$
for ${\bf{A}} \in {\mathbb{R}^{m \times n}}$.



\textit{{Lemma 3:}}
Under the condition and definition in Lemma 2,
if $R'' \neq 0$ on $(0,\infty )$,
then, ${\rm{rank}}\left( {\nabla _{\bf{X}}^2R({{\boldsymbol{\sigma }}_r}({\bf{X}}))} \right) = {r^2}$
and the $r^2$ nonzero eigenvalues of $\nabla _{\bf{X}}^2R({{\boldsymbol{\sigma }}_r}({\bf{X}}))$ are given by
\[{\rm{diag}}({\bf{\Sigma ''}} \otimes {{\bf{I}}_r} + {{\bf{I}}_r} \otimes {\bf{\Sigma ''}}).\]
Further suppose that $R''$ is a nondecreasing function on $(0,\infty )$, then it holds
\[{\bf{0}}\succeq \nabla _{\bf{X}}^2R({{\boldsymbol{\sigma }}_r}({\bf{X}})) \succeq R''({\sigma _r}({\bf{X}})){{\bf{I}}_{mn}}.\]


\subsection{Convergence for A Generalized Nonconvex Penalty}

In the following, let ${{\bf{{ P}}}_\Omega }$ denote the matrix
${{\bf{{ P}}}_\Omega }(i,j) = {\rm{I}}\left( {(i,j) \in \Omega } \right)$,
such that ${{\mathcal{P}}_\Omega }({\bf{X}}) = {{\bf{{ P}}}_\Omega } \odot {\bf{X}}$.
Then, the Hessian of $G$ can be expressed as
\[\nabla _{\bf{X}}^2G({\bf{X}}) = {\bf{diag}}({\rm{vec}}({{\bf{{ P}}}_\Omega })).\]
It is easy to see that ${\lambda _{\max }}\left( {\nabla _{\bf{X}}^2G({\bf{X}})} \right) = 1$.
Then, for a generalized nonconvex penalty satisfying the KL property,
the global convergence of the PGD algorithm to a stationary point can
be directly derived from the results in [37], which is given as follows.

\textit{{Property 2 [37].}} \textit{[Convergence to stationary point]:}
Let $\{ {{\bf{X}}^k}\}$ be a sequence generated by the PGD algorithm (\ref{II-16}),
suppose that $\bar R$ is a closed, proper, lower semi-continuous functions,
if $L > 1$, there hold

\hangindent 2.2em
(i) the sequence $\{ F({{\bf{X}}^k})\} $ is nonincreasing as
\begin{align} \notag
F({{\bf{X}}^{k + 1}}) \le F({{\bf{X}}^k}) - \frac{{L - 1}}{2}\left\| {{{\bf{X}}^{k + 1}} - {{\bf{X}}^k}} \right\|_F^2,
\end{align}
and there exists a constant ${F^*}$ such that $\mathop {\lim }\limits_{k \to \infty } F({{\bf{X}}^k}) = {F^*}$;

\hangindent 2.2em
(ii) ${\left\| {{{\bf{X}}^{k + 1}} - {{\bf{X}}^k}} \right\|_F} \to 0$ as $k \to \infty$,
$\{ {{\bf{X}}^k}\} $ converges to a cluster point set,
and any cluster point is a stationary point of $F$;

\hangindent 2.2em
(iii) further, if there exists a point ${{\bf{X}}^*}$ at which
$F$ satisfies the KL property, $\{ {{\bf{X}}^k}\}$ has finite length
\[\sum\limits_{k = 1}^\infty  {{{\left\| {{{\bf{X}}^{k + 1}} - {{\bf{X}}^k}} \right\|}_F}}  < \infty  \]
and $\{ {{\bf{X}}^k}\} $ converges to ${{\bf{X}}^*}$.

Property 2(i) establishes the sufficient decrease property of the objective $F$, which is a basic property desired for a descent algorithm. Property 2(ii) establishes the subsequence convergence of the PGD algorithm, whilst (iii) establishes the global convergence of the PGD algorithm to a stationary point. Property 2(iii) obviously holds if $\bar R$ is a KL function. The global convergence result applies to a generalized nonconvex penalty $\bar R$ as long as it satisfies the KL property. The KL property is satisfied by most popular nonconvex penalties, such as the hard, $\ell_q$, SCAD and firm thresholding penalties.

\subsection{Convergence for Discontinuous Thresholding}

Among existing nonconvex penalties, there is an important class
which has discontinuous thresholding functions (also referred to
as ``jumping thresholding'' in \cite{48,49,50}), including the popular
${\ell _0}$, ${\ell _q}$, MC, firm thresholding and log-$q$ penalties.
For such penalties, we present more deep analysis
on the convergence properties of the PGD algorithm.

The first result is on the rank
convergence of the sequence $\{ {{\bf{X}}^k}\}$ generated by the PGD algorithm.

\textit{Lemma 4. [Rank convergence]:}
Let $\{ {{\bf{X}}^k}\}$ be a sequence generated by the PGD algorithm (\ref{II-16}).
Suppose that $R$ satisfies Assumption 1 and 2, if $L>1$,
then for any cluster point ${\bf{X}}^*$,
there exist two positive integers ${k}^*$ and $r$ such that, when $k > {k^*}$,
\[{\rm{rank(}}{{\bf{X}}^k}{\rm{)}} = {\rm{rank(}}{{\bf{X}}^*}{\rm{)}} = r.\]

\textit{Proof:} See Appendix C.

This lemma implies that the rank of ${{\bf{X}}^k}$ only changes finitely many times.
By Lemma 4, when $k > {k^*}$, the rank of ${{\bf{X}}^k}$ freezes,
i.e., ${\rm{rank}}({{\bf{X}}^k}) = r$, $\forall k > {k^*}$.
Let ${\bf{X}}$ be a rank-$r$ matrix,
when $k > {k^*}$, minimizing the objective $F$ in (\ref{I-6}) is equivalent to minimizing the following objective
\begin{equation} \label{III-18}
\bar F({\bf{X}}): = \frac{1}{2}\left\| {{{\bf{Y}}_\Omega } - {{ \mathcal{P}}_\Omega }({\bf{X}}))} \right\|_F^2 + \lambda R({{\boldsymbol{\sigma }}_r}({\bf{X}})).
\end{equation}

For $k > {k^*}$, we consider the equivalent objective (\ref{III-18}),
as $\bar F$ is $C^2$ when ${{\boldsymbol{\sigma }}_r}({\bf{X}}) > 0$
(as $R$ is $C^2$ on $(0,\infty )$ by Assumption 1),
which facilitates further convergence analysis of ${\{ {{\bf{X}}^k}\} _{k > {k^*}}}$.
By Lemma 4, the convergence of the whole sequence ${\{ {{\bf{X}}^k}\}}$
is equivalent to the convergence of the sequence ${\{ {{\bf{X}}^k}\} _{k > {k^*}}}$.

Next, we provide a global convergence result for discontinuous thresholding penalties.

\textit{Theorem 1. [Convergence to local minimizer]:}
Under conditions of Lemma 4,
suppose that $R$ is a KL function or satisfies the KL property at a
cluster point of the sequence $\{ {{\bf{X}}^k}\}$,
if $L>1$, then $\{ {{\bf{X}}^k}\}$ converges to a stationary point $ {{\bf{X}}^*}$ of $F$.
Further, let $r{\rm{ = rank}}({{\bf{X}}^*})$, if
\begin{equation} \label{III-19}
\lambda \nabla _{\bf{X}}^2R({{\boldsymbol{\sigma }}_r}({{\bf{X}}^*})) + {\bf{diag}}({\rm{vec}}({{\bf{{ P}}}_\Omega }))\succeq{\bf{0}}
\end{equation}
$ {{\bf{X}}^*}$ is a local minimizer of $F$.

The convergence to a stationary point can be directly claimed from Property 2.
The convergence to a local minimizer is proved in Appendix D.
Let $\sigma  = \min ({{\boldsymbol{\sigma }}_r}({{\bf{X}}^*}))= {{{\sigma }}_r}({{\bf{X}}^*})$,
a sufficient condition for (\ref{III-19}) is
\begin{equation} \label{III-20}
R''(\sigma ) \ge 0.
\end{equation}
This can be justified as follows.
By Lemma 2 and 3, under Assumption 1,
the Hessian of $R({{\boldsymbol{\sigma }}_r}({{\bf{X}}}))$ at ${{\bf{X}}^*}$satisfies
\begin{equation}\notag
\nabla _{\bf{X}}^2R({{\boldsymbol{\sigma }}_r}({{\bf{X}}^*})) \succeq R''(\sigma ){{\bf{I}}_{mn}}
\end{equation}
which together with $\min ({\rm{vec}}({{\bf{{ P}}}_\Omega })) = 0$, for any nonempty
$\Omega  \subset [1, \cdots ,m] \times [1, \cdots ,n]$,
and the Weyl Theorem
implies that the condition (\ref{III-19}) is satisfied if (\ref{III-20}) holds.
Obviously, the sufficient condition (\ref{III-20}) is satisfied
by the hard-thresholding penalty, for which $R''(\sigma )=0$.

\textit{Corollary 1. [Convergence for hard thresholding]:}
Let $\{ {{\bf{X}}^k}\}$ be a sequence generated by the PGD algorithm (16),
$R$ is the hard-thresholding penalty, if $L>1$, $\{ {{\bf{X}}^k}\}$
converges to a local minimizer ${{\bf{X}}^*}$ of $F$.

Next, we show that the nonconvex matrix completion problem (\ref{I-6})
does not have strictly local minimizer,
but has restricted strictly local minimizer.
Specifically, if ${{\bf{X}}^*}$ is a strictly local minimizer
of $F$ with ${\rm{rank}}({{\bf{X}}^*}){\rm{ = }}r$, then
for any sufficiently
small ${\bf{E}} \in {\mathbb{R}^{m \times n}}$ satisfying
${\rm{rank}}({{\bf{X}}^*} + {\bf{E}}){{ = }}r$,
it holds $\bar F({{\bf{X}}^*} + {\bf{E}}) > \bar F({{\bf{X}}^*})$,
hence $\nabla _{\bf{X}}^2\bar F({{\bf{X}}^*})\succ{\bf{0}}$.
However, when $r<\min(m,n)$,
${\lambda _{\max }}\left( {\nabla _{\bf{X}}^2R({{\boldsymbol{\sigma }}_r}({{\bf{X}}^*}))} \right) = {{0}}$
by Assumption 1 and Lemma 3,
which together with ${\lambda _{\min }}\left( {{\bf{diag}}({\rm{vec}}({{\bf{{ P}}}_\Omega }))} \right) = {{0}}$
and the Weyl Theorem implies that
\[{\lambda _{\min }}\left( {\nabla _{\bf{X}}^2\bar F({{\bf{X}}^*})} \right) \le 0.\]
That is $\nabla _{\bf{X}}^2\bar F({{\bf{X}}^*})$ cannot be positive definite.
Thus, ${{\bf{X}}^*}$ cannot be a strictly local minimizer of $F$,
and the strictly local minimizer set of $F$ is empty.
Despite of this, we have the following result of convergence to a
restricted strictly local minimizer. In the following,
let $\nabla _{{{\bf{X}}_\Omega }}^2R$ denote the submatrix of
$\nabla _{\bf{X}}^2R$ corresponding to the index subset $\Omega $.

\textit{Theorem 2. [Convergence to $\Omega $-restricted strictly local minimizer]:}
Under conditions of Lemma 4, suppose that $R$ is a KL function or
satisfies the KL property at a cluster point of the sequence $\{ {{\bf{X}}^k}\} $,
then $\{ {{\bf{X}}^k}\} $ converges to a stationary point ${{\bf{X}}^*}$ of $F$.
Further, let $r{\rm{ = rank}}({{\bf{X}}^*})$, if
\begin{equation}\label{III-21}
\lambda \nabla _{{{\bf{X}}_\Omega }}^2R({{\boldsymbol{\sigma }}_r}({{\bf{X}}^*})) + {{\bf{I}}_{|\Omega |}} \succ {\bf{0}}
\end{equation}
${{\bf{X}}^*}$ is a $\Omega $-restricted strictly local minimizer of $F$.

The proof is given in Appendix E. Since
$\nabla _{\bf{X}}^2R({{\boldsymbol{\sigma }}_r}({{\bf{X}}^*}))\succeq R''(\sigma ){{\bf{I}}_{mn}}$,
it is easy to see that
\[\nabla _{{{\bf{X}}_\Omega }}^2R({{\boldsymbol{\sigma }}_r}({{\bf{X}}^*}))\succeq R''(\sigma ){{\bf{I}}_{|\Omega|}}.\]
Then, the condition in (\ref{III-21}) is equivalent to
\begin{equation}\label{III-22}
1 + \lambda R''(\sigma ) > 0.
\end{equation}
By this Theorem, we have the following result for the ${\ell _q}$ ($0 < q < 1$) penalty.

\textit{Corollary 2. [Convergence for ${\ell _q}$ penalty]:}
Let $\{ {{\bf{X}}^k}\} $ be a sequence generated by the PGD algorithm (\ref{II-16}),
$R$ is the ${\ell _q}$ penalty with $0 < q < 1$, if $L>1$,
$\{ {{\bf{X}}^k}\} $ converges to a stationary point ${ {{\bf{X}}^*}} $ of $F$.
Further, if
\begin{equation}\label{III-23}
\lambda  < \frac{{{\sigma ^{2 - q}}}}{{q(1 - q)}}~~~~{\rm{or}}~~~~L < \frac{2}{q}
\end{equation}
then ${ {{\bf{X}}^*}} $ is a $\Omega $-restricted strictly local minimizer of $F$.

For the ${\ell _q}$ ($0 < q < 1$) penalty,
\[R''(\sigma ) = q(q - 1){\sigma ^{q - 2}}\]
which together with (\ref{III-22}) results in the left hand of (\ref{III-23}).
The right hand condition in (\ref{III-23}) follows from the property of the $\ell_q$-thresholding (see Table I) and (\ref{II-16}) that
\begin{equation}\notag
\sigma  = \min ({{\boldsymbol{\sigma }}_r}({{\bf{X}}^*})) \ge {\left( {\frac{{2(1 - q)\lambda }}{L}} \right)^{\frac{1}{{2 - q}}}}.
\end{equation}

Furthermore, for the hard-thresholding penalty,
the convergence to a $\Omega $-restricted strictly local minimizer is straightforward if $L>1$.

\subsection{Eventually Linear Convergence Rate for Discontinuous Thresholding}

This subsection derives the linear convergence of the
PGD algorithm for nonconvex penalties with discontinuous
thresholding function. Before proceeding to the analysis,
we first show some properties on the sequence
$\{ {{\bf{X}}^k}\}$  in the neighborhood of ${{\bf{X}}^*}$.

Consider a neighborhood of ${{\bf{X}}^*}$ as
\begin{align}\notag
\mathcal{N}({{\bf{X}}^*},\delta ) = &\{{\bf{X}} \in {\mathbb{R}^{m \times n}}:\\ \notag
&{\left\| {{\bf{X}} - {{\bf{X}}^*}} \right\|_F} < \delta ,{\rm{rank}}({\bf{X}}) = {\rm{rank}}({{\bf{X}}^*}) = r\}
\end{align}
for any $0 < \delta  < {\beta _L}$, ${\beta _L}$ is the
``jumping'' size of the thresholding function
${P_{ R,L/\lambda }}$ (corresponding to ${\bar P_{\bar R,L/\lambda }}$ in (\ref{II-16}))
at the its threshold point.
Under Assumption 1, $\nabla _{\bf{X}}^2R({{\boldsymbol{\sigma }}_r}({{\bf{X}}^*})) \ge R''(\sigma ){{\bf{I}}_{mn}}$ by Lemma 3
and $R''$ is nondecreasing on $(0, + \infty )$, thus,
there exists a sufficiently small constant ${c_R} > 0$,
which is dependent on $\delta $ and ${c_R} \to 0$ as $\delta  \to 0$, such that.
\begin{equation}\label{III-25}
\begin{split}
&\left\langle {{\nabla _{\bf{X}}}R({{\boldsymbol{\sigma }}_{{r}}}({\bf{X}})) - {\nabla _{\bf{X}}}R({{\boldsymbol{\sigma }}_{{r}}}({{\bf{X}}^*})),{\bf{X}} - {{\bf{X}}^*}} \right\rangle \\
& \ge (R''(\sigma ) - {c_R})\left\| {{\bf{X}} - {{\bf{X}}^*}} \right\|_F^2.
\end{split}
\end{equation}

For the second property, we denote ${\bf{Q}} = {\bf{X}} - \frac{1}{L}\left[ {{{\mathcal{P}}_\Omega }({\bf{X}}) - {{\bf{Y}}_\Omega }} \right]$ and ${{\bf{Q}}^*} = {{\bf{X}}^*} - \frac{1}{L}\left[ {{{ \mathcal{P}}_\Omega }({{\bf{X}}^*}) - {{\bf{Y}}_\Omega }} \right]$ for some $L > 1$, which have the following full SVD
\[{\bf{Q}} = [{\bf{U}},{{\bf{U}}_ \bot }]\left[ {\begin{array}{*{20}{c}}
\!{\bf{\Sigma }}&{\bf{0}}\!\\
\!{\bf{0}}&{{{\bf{\Sigma }}_ \bot }}\!
\end{array}} \right]{[{\bf{V}},{{\bf{V}}_ \bot }]^T}\]
\[{{\bf{Q}}^*} = [{{\bf{U}}^*},{\bf{U}}_ \bot ^*]\left[ {\begin{array}{*{20}{c}}
\!{{{\bf{\Sigma }}^*}}&{\bf{0}}\!\\
\!{\bf{0}}&{{\bf{\Sigma }}_ \bot ^*}\!
\end{array}} \right]{[{{\bf{V}}^*},{\bf{V}}_ \bot ^*]^T}\]
where ${\bf{U}},{{\bf{U}}^*} \in {\mathbb{R}^{m \times r}}$, ${\bf{V}},{{\bf{V}}^*} \in {\mathbb{R}^{n \times r}}$ and
\[{\bf{\Sigma }} = {\bf{diag}}({{\boldsymbol{\sigma }}_r}({\bf{Q}})),~~~~{{\bf{\Sigma }}_ \bot } = {\bf{diag}}({{\boldsymbol{\sigma }}_{r \bot }}({\bf{Q}}))\]
\[{{\bf{\Sigma }}^*} = {\bf{diag}}({{\boldsymbol{\sigma }}_r}({{\bf{Q}}^*})),~~~~{\bf{\Sigma }}_ \bot ^* = {\bf{diag}}({{\boldsymbol{\sigma }}_{r \bot }}({{\bf{Q}}^*})).\]
Let
\[{{\bf{Q}}_r} = {\bf{U\Sigma }}{{\bf{V}}^T},~~~~{{\bf{Q}}_{r \bot }} = {{\bf{U}}_ \bot }{{\bf{\Sigma }}_ \bot }{\bf{V}}_ \bot ^T\]
\[{\bf{Q}}_r^* = {{\bf{U}}^*}{{\bf{\Sigma }}^*}{{\bf{V}}^*}^T,~~~{\bf{Q}}_{r \bot }^* = {\bf{U}}_ \bot ^*{\bf{\Sigma }}_ \bot ^*{\bf{V}}_ \bot ^{*T}.\]
Then, it follows that ${\bf{Q}} = {{\bf{Q}}_r} + {{\bf{Q}}_{r \bot }}$,
${{\bf{Q}}^*} = {\bf{Q}}_r^* + {\bf{Q}}_{r \bot }^*$ and
\[\left\| {{\bf{Q}} - {{\bf{Q}}^*}} \right\|_F^2 = \left\| {{{\bf{H}}_1}} \right\|_F^2 + \left\| {{{\bf{H}}_2}} \right\|_F^2 + 2\left\langle {{{\bf{H}}_1},{{\bf{H}}_2}} \right\rangle \]
where ${{\bf{H}}_1} = {{\bf{Q}}_r} - {\bf{Q}}_r^*$ and
${{\bf{H}}_2} = {{\bf{Q}}_{r \bot }} - {\bf{Q}}_{r \bot }^*$.
When $\delta  \to 0$ (hence ${\left\| {{\bf{X}} - {{\bf{X}}^*}} \right\|_F} \to 0$ and ${\left\| {{\bf{Q}} - {{\bf{Q}}^*}} \right\|_F} \to 0$), the range space of ${{\bf{H}}_1}$, denoted by $\mathcal{R}({{\bf{H}}_1})$, tends to be orthogonal with the range space of ${{\bf{H}}_2}$, denoted by $\mathcal{R}({{\bf{H}}_2})$. In other words, let ${\boldsymbol{\theta }}(\mathcal{R}({{\bf{H}}_1}),\mathcal{R}({{\bf{H}}_2}))$ be a vector contains the principal angles between the two range spaces $\mathcal{R}({{\bf{H}}_1})$ and $\mathcal{R}({{\bf{H}}_2})$, it follows that
\[{\left\| {\cos {\boldsymbol{\theta }}(\mathcal{R}({{\bf{H}}_1}),\mathcal{R}({{\bf{H}}_2}))} \right\|_2} \to 0 ~~{\rm{as}}~~ \delta  \to 0.\]
Based on this fact, for each ${\bf{X}} \in \mathcal{N}({{\bf{X}}^*},\delta )$
there exists a constant $\alpha ({\bf{X}}) \in [- \frac{1}{2},\frac{1}{2}]$
which is dependent on $\delta $, satisfying $\alpha ({\bf{X}}) \to 0$ as $\delta  \to 0$, such that
\begin{equation}\label{III-26}
\left\langle {{{\bf{H}}_1},{{\bf{H}}_2}} \right\rangle  = \alpha ({\bf{X}})\big( {\left\| {{{\bf{H}}_1}} \right\|_F^2 + \left\| {{{\bf{H}}_2}} \right\|_F^2} \big).
\end{equation}

For any ${\bf{X}} \in \mathcal{N}({{\bf{X}}^*},\delta )$, when ${{\bf{X}}^*}$ is a stationary point of the $F$ (hence a fixed point of the PGD algorithm, i.e., ${{\bf{X}}^*} = {\bar P_{\bar R,L/\lambda }}({{\bf{Q}}^*}) = {\bar P_{\bar R,L/\lambda }}({\bf{Q}}_r^*)$), it holds ${\left\| {{{\bf{Q}}_r} - {\bf{Q}}_r^*} \right\|_F} > 0$ if ${\bf{X}} \ne {{\bf{X}}^*}$, since ${{\bf{X}}^*} \ne {\bar P_{\bar R,L/\lambda }}({\bf{Q}}) = {\bar P_{\bar R,L/\lambda }}({{\bf{Q}}_r})$ in this case. Meanwhile, a basic assumption which makes the matrix completion problem meaningful is that, the underlying low-rank matrix ${\bf{M}}$ is generated from a random orthogonal model (hence not sparse), whilst the cardinality is sampled uniformly at random [1], [2]. Based on these assumptions we can reasonably further make the following assumption.

\textit{Assumption 3:}
For ${\bf{X}} \in \mathcal{N}({{\bf{X}}^*},\delta )$ with a sufficiently small $\delta$
(hence $\alpha ({\bf{X}})$ in (\ref{III-26}) is sufficiently small),
\[\left\| {{{\bf{Q}}_{r \bot }} - {\bf{Q}}_{r \bot }^*} \right\|_F^2 = \gamma ({\bf{X}})\left\| {{\bf{Q}} - {{\bf{Q}}^*}} \right\|_F^2\]
\[\left\| {{{\mathcal{P}}_\Omega }({\bf{X}} - {{\bf{X}}^*})} \right\|_F^2 = \xi ({\bf{X}})\left\| {{\bf{X}} - {{\bf{X}}^*}} \right\|_F^2\]
\[\left\| {{\mathcal{P}}_\Omega ^ \bot ({\bf{X}} - {{\bf{X}}^*})} \right\|_F^2 = \big(1 - \xi ({\bf{X}})\big)\left\| {{\bf{X}} - {{\bf{X}}^*}} \right\|_F^2\]
for some $\gamma ({\bf{X}}) \in [0,1)$ and $\xi ({\bf{X}}) \in (0,1)$,
with $\gamma ({\bf{X}})$ and $\xi ({\bf{X}})$ be respectively
lower bounded by $\gamma  \in [0,1)$ and $\xi  \in (0,1)$.
Meanwhile, $\alpha ({\bf{X}}) = 0$ if $\gamma ({\bf{X}}) = 0$
(since ${\left\| {{{\bf{Q}}_r} - {\bf{Q}}_r^*} \right\|_F} > 0$ if ${\bf{X}} \ne {{\bf{X}}^*}$).

With the above properties, we obtain the following result.

\textit{Theorem 3. [Eventually linear rate for discontinuous thresholding]:}
Under conditions of Theorem 2 and Assumption 3, if
\[1 + \lambda R''(\sigma ){\rm{/}}L > \sqrt {(1 - \gamma )(1 - 2\xi {\rm{/}}L + \xi {\rm{/}}{L^2})} \]
then $\{ {{\bf{X}}^k}\} $ converges to a stationary point ${{\bf{X}}^*}$
of $F$ with an eventually linear convergence rate, i.e.,
there exists a positive integer ${k^0}$ and a constant $\rho  \in (0,1)$
such that when $k > {k^0}$,
\[{\left\| {{{\bf{X}}^{k + 1}} - {{\bf{X}}^*}} \right\|_F} \le \rho {\left\| {{{\bf{X}}^k} - {{\bf{X}}^*}} \right\|_F}\]
\[{\left\| {{{\bf{X}}^{k + 1}} - {{\bf{X}}^*}} \right\|_F} \le \frac{\rho }{{1 - \rho }}{\left\| {{{\bf{X}}^{k + 1}} - {{\bf{X}}^k}} \right\|_F}.\]

The proof is given in Appendix F.
For the matrix completion problem,
the range space convergence property (25) and
the nondegenerate conditions in Assumption 3 are needed
to derive the local linear convergence for the
singular-value thresholding based PGD algorithm.
Based on this Theorem, we have the following result for the ${\ell _q}$ penalty.

\textit{Corollary 3. [Eventually linear rate for ${\ell _q}$ penalty]:}
Under conditions of Corollary 2 and Assumption 3, if
\[1 + \lambda q(q - 1){\sigma ^{q - 2}}{\rm{/}}L > \sqrt {(1 - \gamma )(1 - 2\xi {\rm{/}}L + \xi {\rm{/}}{L^2})} \]
then $\{ {{\bf{X}}^k}\} $ converges to a stationary point ${{\bf{X}}^*}$ (also a $\Omega$-restricted strictly local minimizer)
of $F$ with an eventually linear convergence rate.

For the hard-thresholding penalty,
eventually linear convergence is more straightforward.

\textit{Corollary 4. [Eventually linear rate for hard thresholding]:}
Under conditions of Corollary 1 and Assumption 3,
$\{ {{\bf{X}}^k}\} $ converges to a local minimizer ${{\bf{X}}^*}$  (also a $\Omega$-restricted strictly local minimizer)
of $F$ with an eventually linear convergence rate.

\section{Numerical Experiments}

\begin{figure}[!t]
 \centering
 \includegraphics[scale = 0.2]{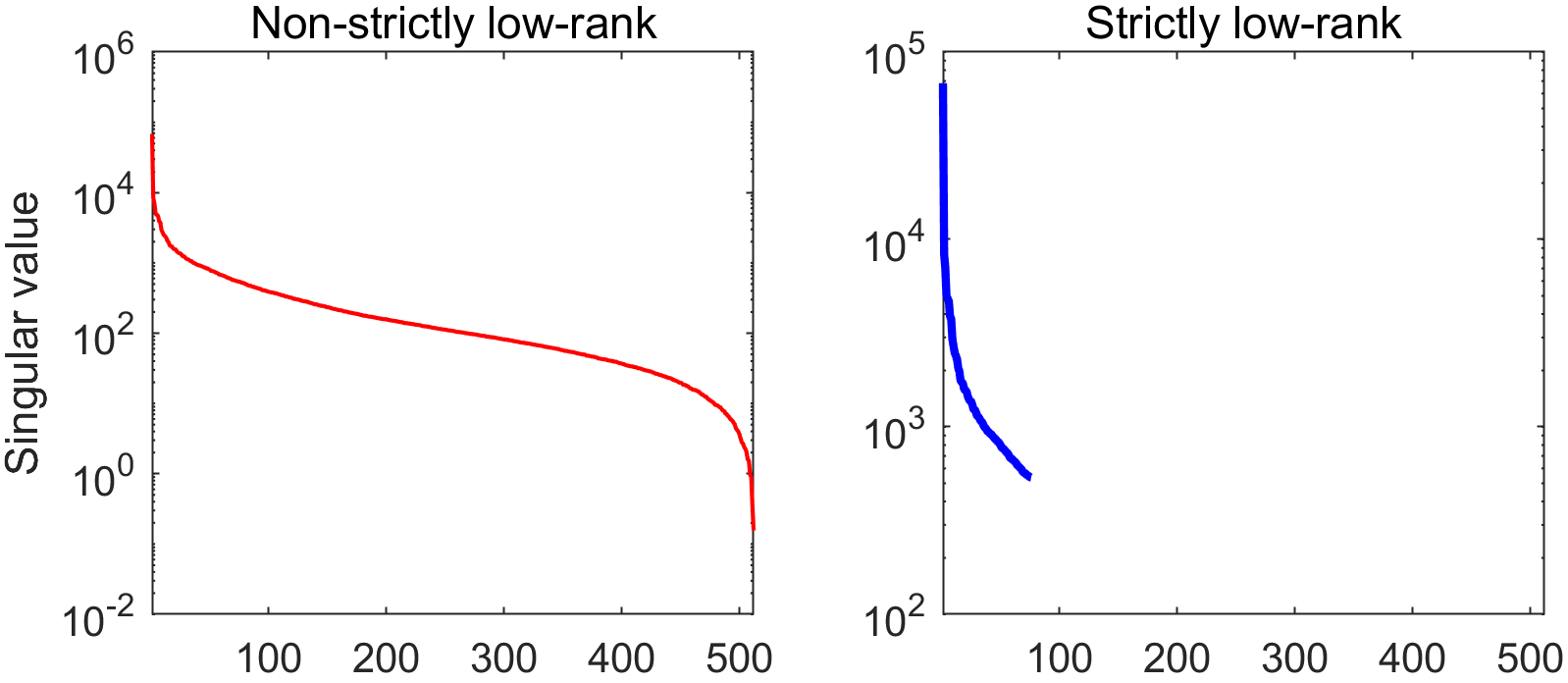}
\caption{Sorted singular values in the two cases. \textit{Left:} Non-strictly low-rank (the original image is used). \textit{Right:} Strictly low-rank (the singular values of the original image are truncated to retain only the largest 15\%).}
 \label{figure2}
\end{figure}

\begin{figure*}[!t]
\centering
\subfigure[Zero initialization]{
{\includegraphics[scale = 0.7]{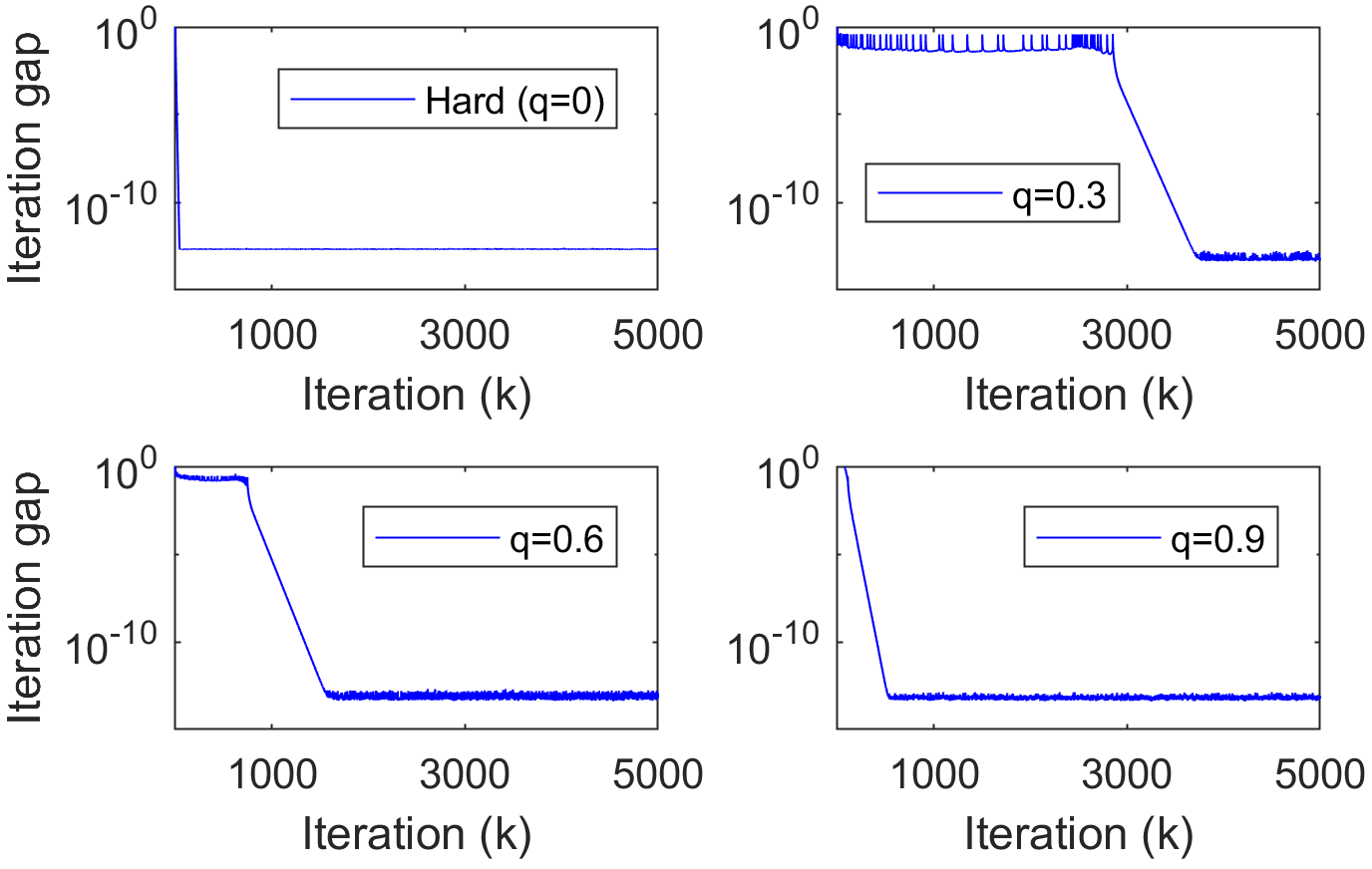}}}~~~~~~
\subfigure[Initialized with the solution of soft-thresholding]{
{\includegraphics[scale = 0.7]{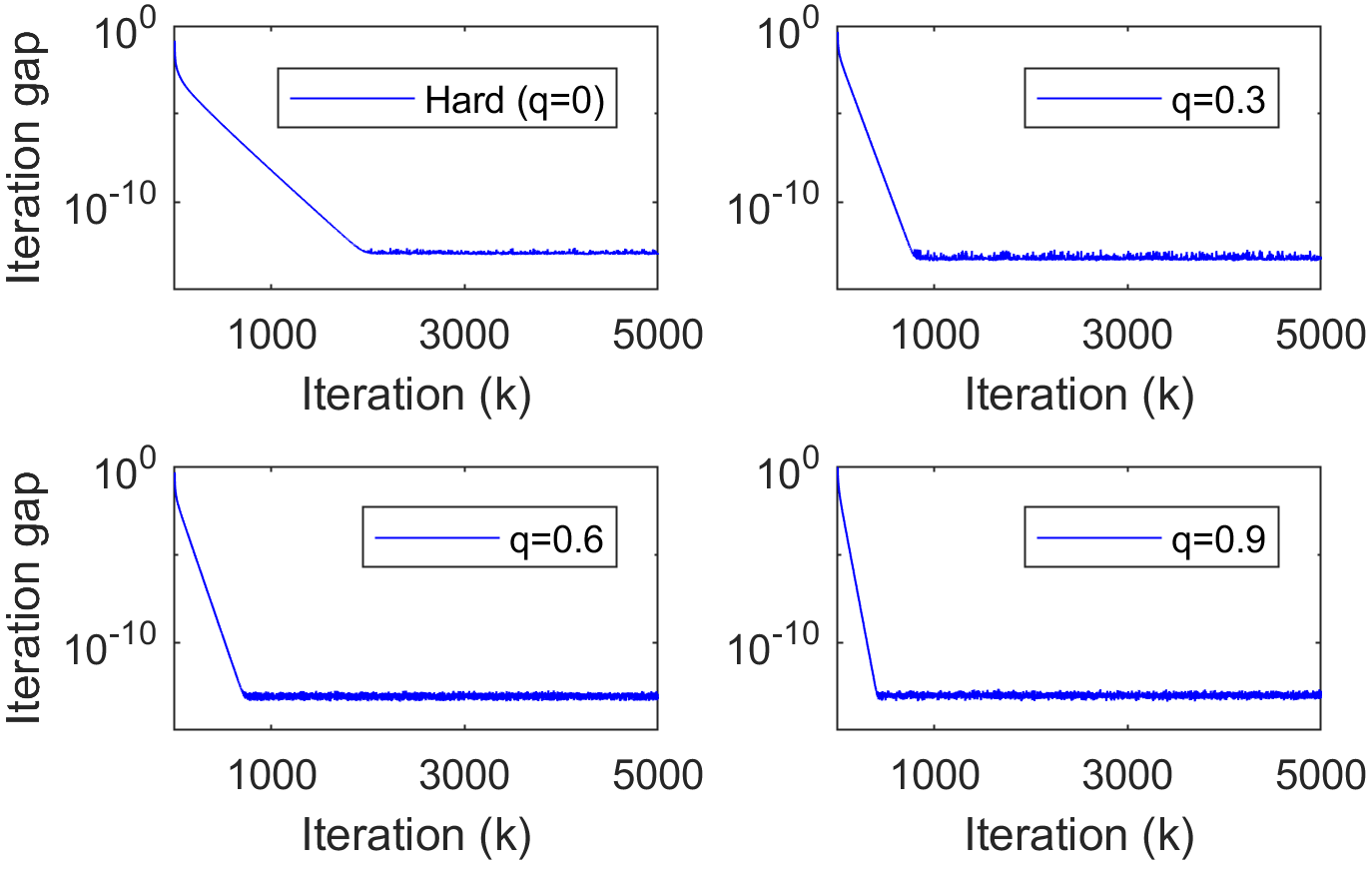}}}
\caption{Convergence behavior of the PGD algorithm in the nonconvex case, (a) initialized with zero, (b) initialized with the solution of soft-thresholding.}
\label{figure3}
\end{figure*}

\begin{figure*}[!t]
\centering
\subfigure[Non-strictly low-rank]{
{\includegraphics[scale = 0.65]{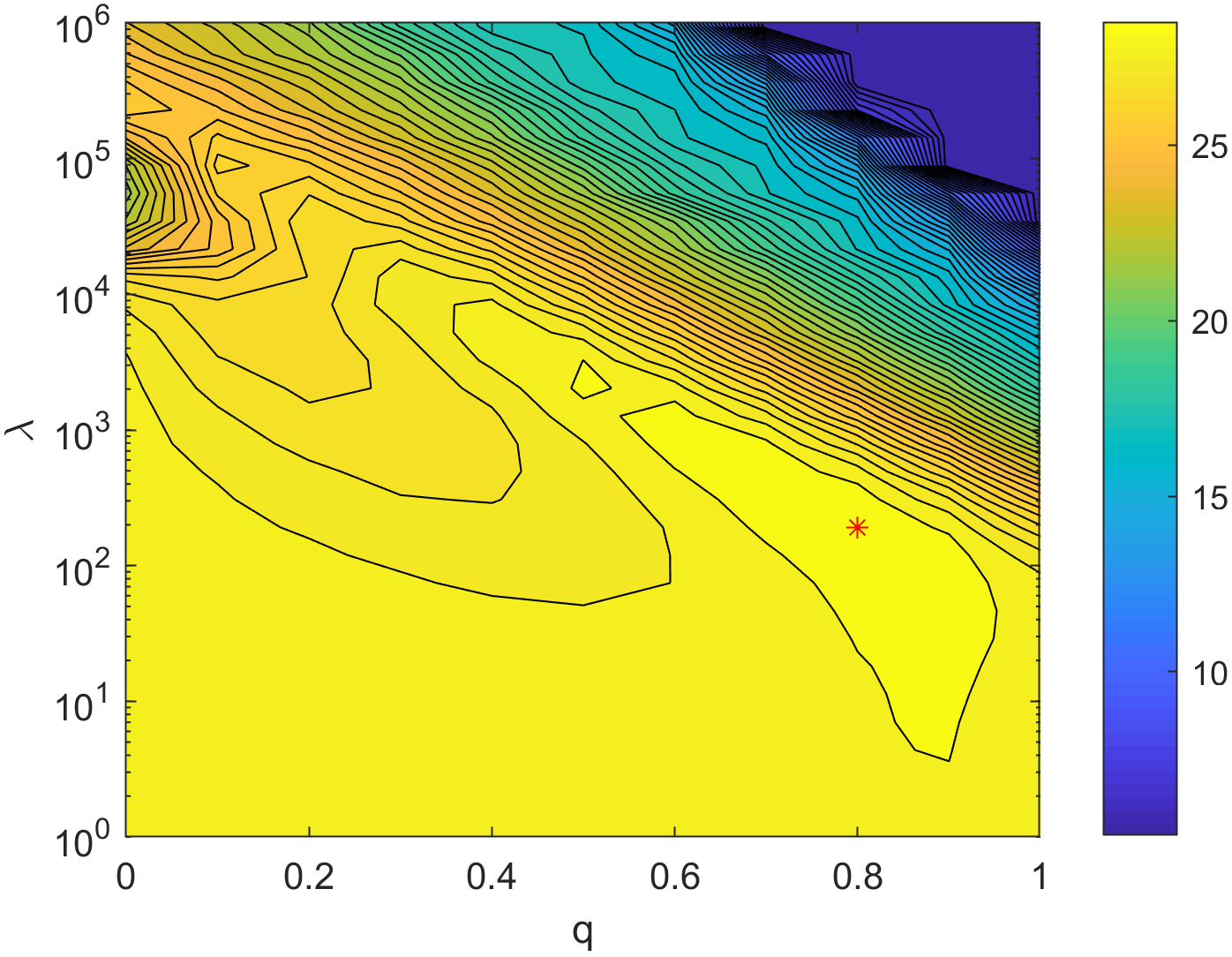}}}~~~~~~~~~~
\subfigure[Strictly low-rank]{
{\includegraphics[scale = 0.65]{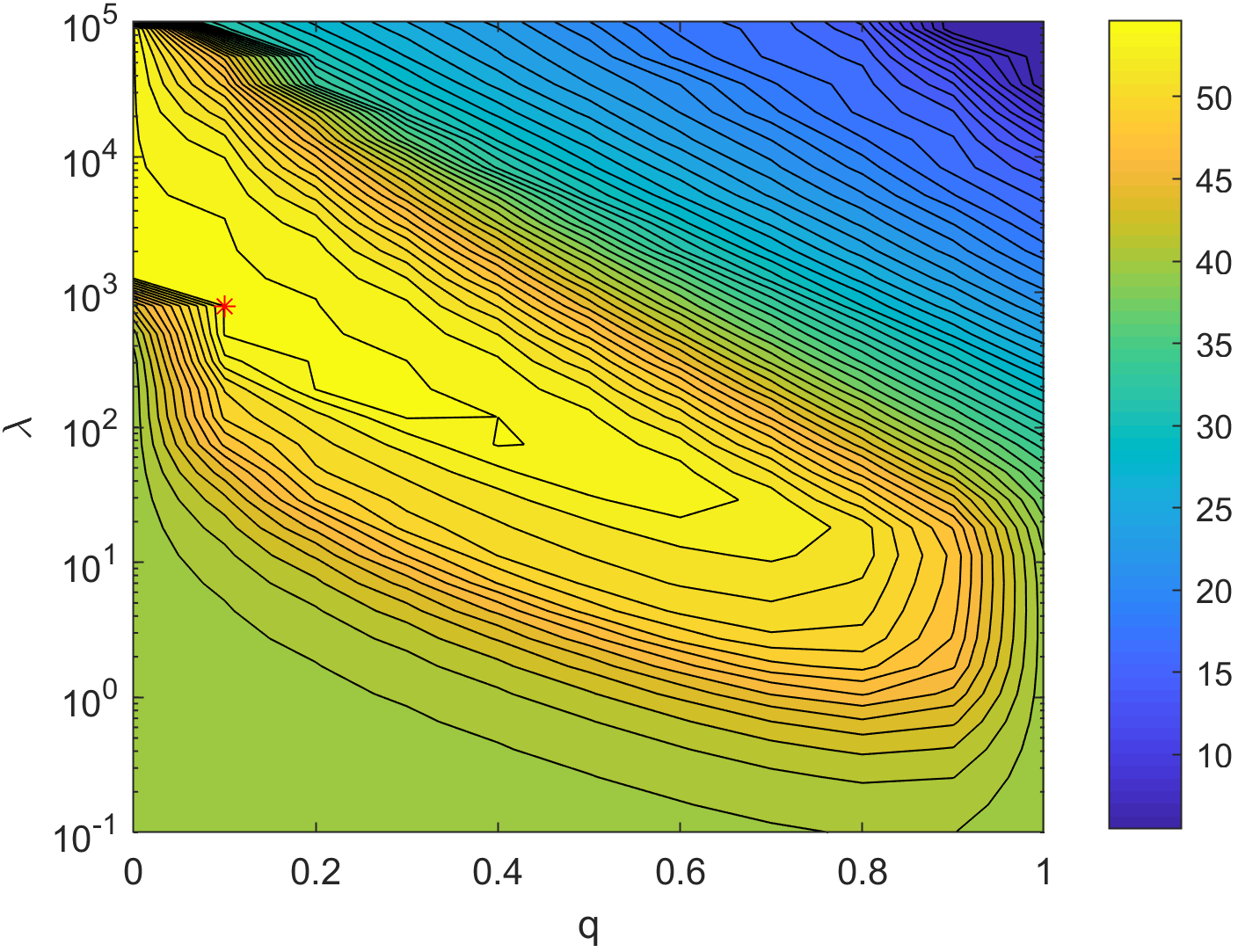}}}
\caption{Recovery PSNR of the PGD algorithm with $\ell_q$ penalty in the case of SNR = 40 dB, (a) non-strictly low-rank, (b) strictly low-rank.}
\label{figure4}
\end{figure*}

\begin{figure*}[!t]
\centering
\subfigure[Non-strictly low-rank]{
{\includegraphics[scale = 0.65]{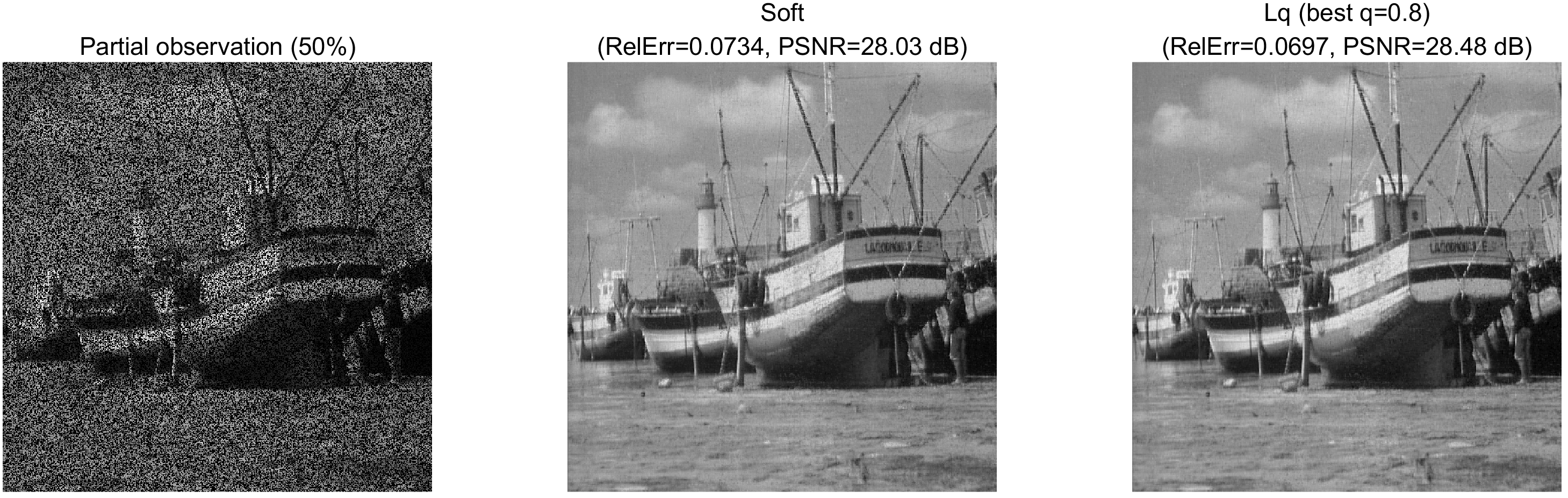}}}\\
\subfigure[Strictly low-rank]{
{\includegraphics[scale = 0.65]{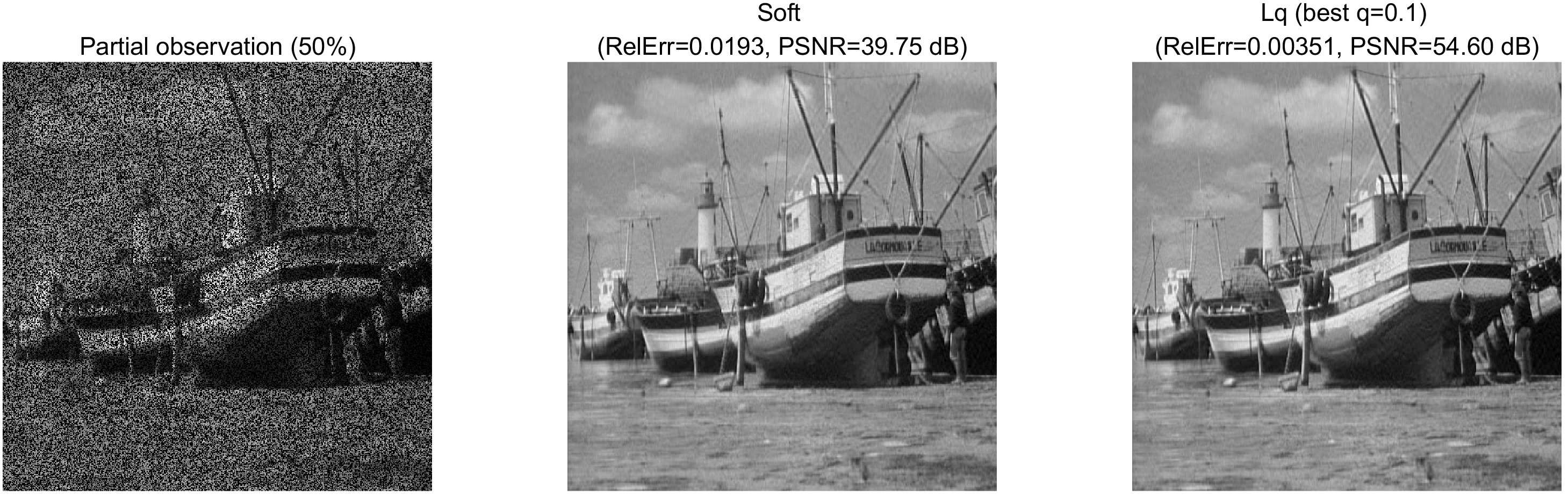}}}
\caption{Recovery performance of the PGD algorithm on strictly and non-strictly low-rank images for SNR = 40 dB (along with the RelErr and PSNR of recovery). The values of $q$ providing the best performance of the $\ell_q$ penalty is presented.}
\label{figure5}
\end{figure*}

\begin{figure*}[!t]
\centering
\subfigure[Non-strictly low-rank]{
{\includegraphics[scale = 0.65]{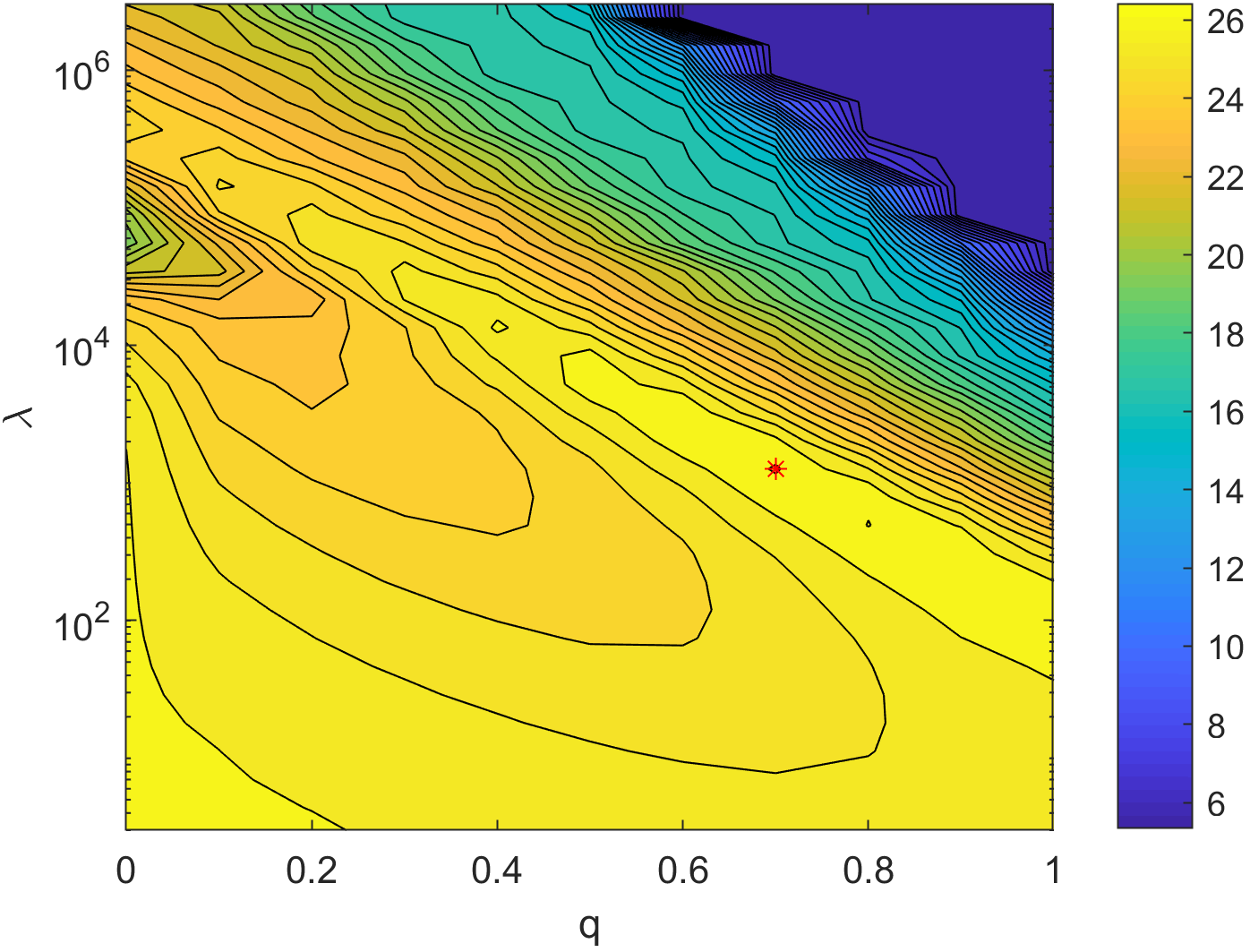}}}~~~~~~~~~~
\subfigure[Strictly low-rank]{
{\includegraphics[scale = 0.65]{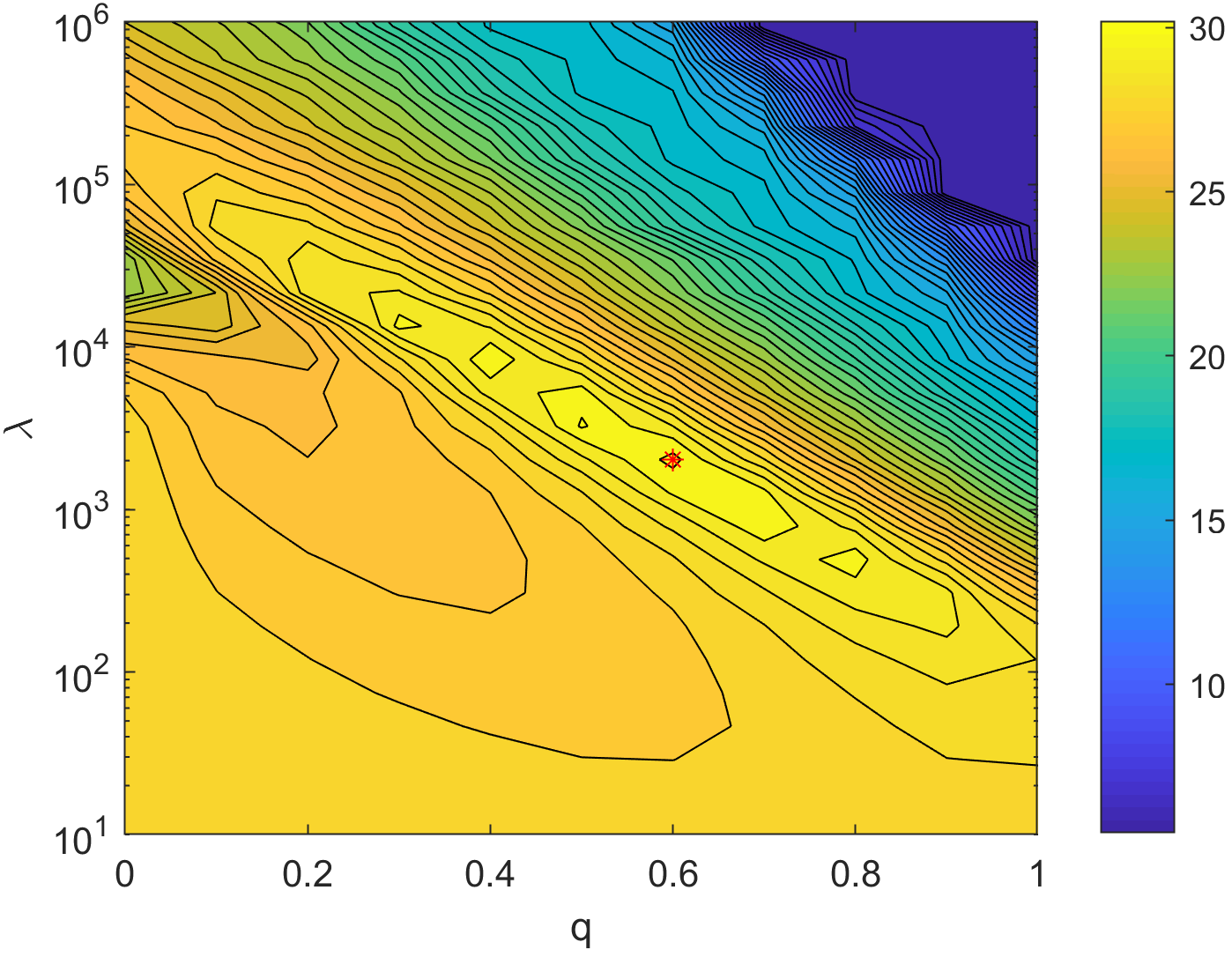}}}
\caption{Recovery PSNR of the PGD algorithm with $\ell_q$ penalty in the case of SNR = 15 dB, (a) non-strictly low-rank, (b) strictly low-rank.}
\label{figure6}
\end{figure*}

\begin{figure*}[!t]
\centering
\subfigure[Non-strictly low-rank]{
{\includegraphics[scale = 0.65]{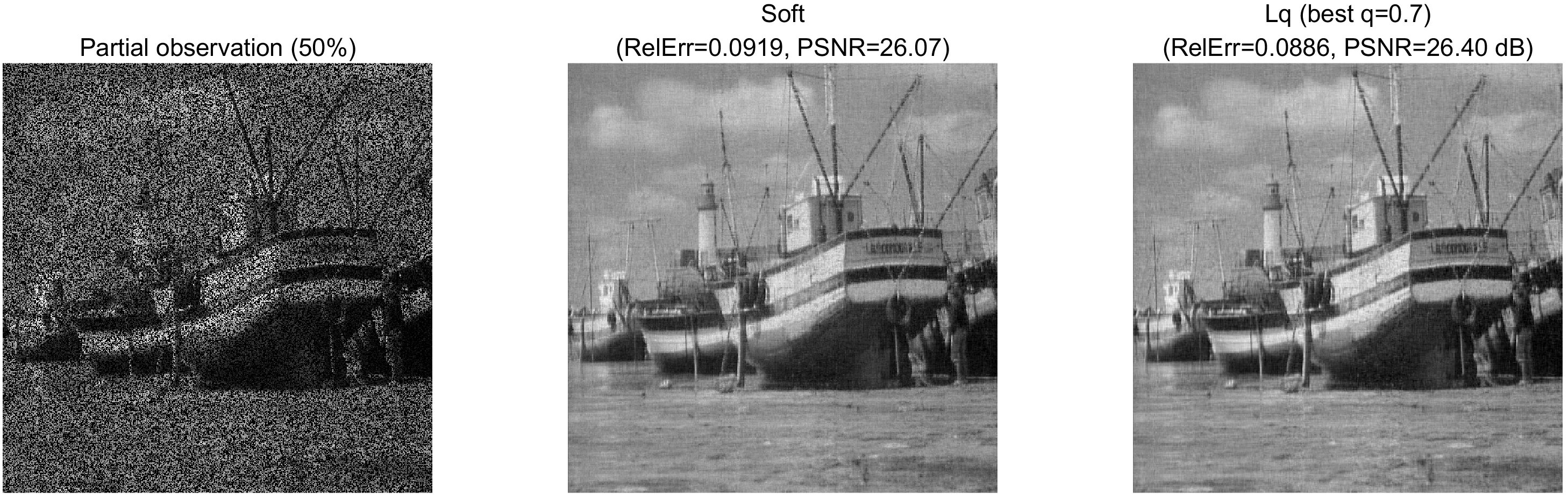}}}\\
\subfigure[Strictly low-rank]{
{\includegraphics[scale = 0.65]{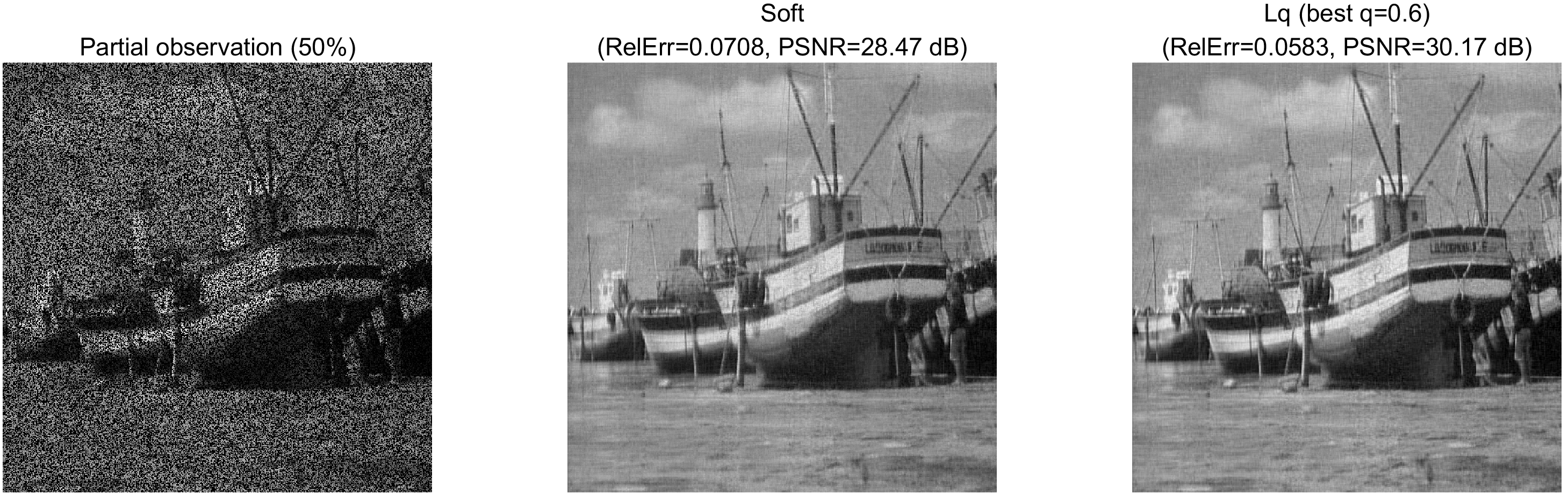}}}
\caption{Recovery performance of the PGD algorithm on strictly and non-strictly low-rank images for SNR = 15 dB (along with the RelErr and PSNR of recovery). The values of $q$ providing the best performance of the $\ell_q$ penalty is presented.}
\label{figure7}
\end{figure*}

In this section, we illustrate the PGD algorithm via
numerical experiments on inpainting. We consider the
$\ell_q$ penalty ($\bar R$ be the Schatten-$q$ norm)
as it has a flexible parametric form that adapts to
different penalty functions by varying the value of $q$.
The goal is to recover a $512\times512$ image from 50\% of
the pixels in the presence of entry noise, which is the case
in many image inpainting and denoising applications
(e.g., the other 50\% of the pixels are corrupted by
salt-and-pepper noise). Two cases are considered:
\textit{1)} Non-strictly low-rank: the original image
is used, which is not strictly low-rank but rather with
singular values approximately following an exponential decay;
\textit{2)} Strictly low-rank: the singular values of the
original image are truncated and only the 15\% largest
values are retained, which results in a strictly low-rank
image used for evaluation.
Fig. 2 plots the sorted singular values in the two cases.

Fig. 3 shows the typical convergence behavior of the PGD
algorithm for $q = \{0,0.3,0.6,0.9\}$ in two initialization
conditions. The iteration gap
${\left\| {{{\bf{X}}^{k + 1}} - {{\bf{X}}^k}} \right\|_F}/\sqrt {mn}$ is plotted.
The results indicate that a good initialization facilitates
the convergence of the PGD algorithm in the nonconvex case.
Meanwhile, with zero initialization, the hard-thresholding
seems to converge to a near local minimizer quickly. Eventually
linear convergence rate of the PGD algorithm with $\ell_q$
penalty can be observed from the iteration gap variation.
As well as most nonconvex algorithms, the performance of
the PGD algorithm is closely related to the initialization.
In the following, for the nonconvex case of $0 \le q < 1$,
we first run the PGD algorithm with $\ell_1$ (nuclear norm) penalty to obtain
an initialization.

Fig. 4 shows the recovery peak-signal noise ratio (PSNR)
of the PGD algorithm for different combinations of $q$
and $\lambda$ in the two considered cases, with entry-wise
Gaussian noise of 40 dB.
Fig. 5 shows the recovered images along
with the relative error of recovery (RelErr)
and PSNR of each recovered image.
Fig. 6 and Fig. 7 show the results for a
higher noise condition with entry-wise noise of 15 dB.
The recovery PSNR comparison between the $\ell_q$ and
$\ell_1$ penalties is provided in Table II.
It can be seen that with a properly selected value of $q$,
the $\ell_q$ penalty outperforms the $\ell_1$ penalty in all cases.
The advantage of the $\ell_q$ penalty over the $\ell_1$ penalty
is more prominent in the strictly low-rank case. For example
in the low noise case with SNR = 40 dB, the advantage in the
strictly low-rank case is about 14.85 dB, while that in the
non-strictly low-rank case is only about 0.45 dB.
This advantage wakens in the high noise case with SNR = 15 dB.

\begin{table}[!t]
\caption{Recovery PSNR comparison (in dB) (along with the values of $q$ providing the best performance of the $\ell_q$ penalty).}
\centering
\newcommand{\tabincell}[2]{\begin{tabular}{@{}#1@{}}#2\end{tabular}}
\begin{tabular}{|c|c|c|c|c|}
\hline
\multirow{2}{*}{ }& \multicolumn{2}{|c|}{SNR = 40 dB}  & \multicolumn{2}{|c|}{SNR = 15 dB} \\
\cline{2-5}
 & $\ell_1$ & {$\ell_q$}  & $\ell_1$ & {$\ell_q$}\\
 \hline
\tabincell{c}{non-strictly\\ low-rank}  & {28.03} &  \tabincell{c}{28.48 \\($q=0.8$)}  & {26.07}  &  \tabincell{c}{26.40 \\($q=0.7$)}\\
\hline
\tabincell{c}{strictly \\low-rank}      &  {39.75} &  \tabincell{c}{54.60 \\($q=0.1$)}  & {28.47} &  \tabincell{c}{30.17 \\($q=0.6$)}\\
\hline
\end{tabular}
\end{table}

Moreover, the results imply that for the $\ell_q$ penalty,
in the low noise condition, e.g., SNR = 40 dB,
a relatively small value of $q$, e.g., $q<0.5$,
should be used in the strictly low-rank case,
while a relatively large value of $q$, e.g., $q>0.5$,
should be used in the non-strictly low-rank case.
However, in the high noise case, e.g., SNR = 15 dB,
a moderate value of $q$ tends to yield good performance.

\section{Conclusion}

This work provided an analysis on the PGD algorithm
for matrix completion using a nonconvex penalty. First, some
properties on the gradient and Hessian of a generalized low-rank
penalty have been established. Then,
we provide more deep analysis on a
popular class of nonconvex penalties which have discontinuous
thresholding functions. For such penalties, we established the
finite rank change, convergence to a restricted strictly local
minimizer and an eventually linear convergence rate for the
PGD algorithm under certain conditions. Meanwhile, convergence
to a local minimizer has been obtained for the PGD algorithm
with hard-thresholding penalty. Experimental results on inpainting
demonstrated that, the benefit of using a nonconvex penalty is
especially conspicuous in recovering a strictly low-rank matrix
in the presence of small noise.

\appendices

\section{Gradient and Hessian of Functions Contains Low-rank Penalty}
In general, a low-rank penalty function is not differential
with respective to a low-rank matrix. For example,
for a generalized low-rank penalty defined as (\ref{II-8}),
$\bar R({\bf{X}}) = R({\boldsymbol{\sigma }}({\bf{X}}))$
for a matrix ${\bf{X}} \in {\mathbb{R}^{m \times n}}$,
since $R$ is usually nonsmooth at zero (such as the penalties mentioned in section II),
$R({\boldsymbol{\sigma }}({\bf{X}}))$ is not differential when ${\rm{rank}}({\bf{X}}) < \min (m,n)$.
However, when $R$ is $C^2$ on $(0, + \infty )$, it is differential
on $C^2$ arcs $t \to {\bf{X}}(t)$ if ${\rm{rank}}({\bf{X}}(t))$ is constant,
although the rank may be less than $\min (m,n)$).
Consider the latter case, we can analytically derive
the gradient and Hessian of a function which contains
a low-rank penalty as a term.

Suppose that ${\bf{X}}$ is of rank $r$, $r \le \min (m,n)$,
with any truncated SVD ${\bf{X}} = {\bf{U\Sigma V}}^T$,
where ${\bf{\Sigma }} = {\bf{diag}}({{\boldsymbol{\sigma }}_r}({\bf{X}}))$,
${\bf{U}} \in {\mathbb{R}^{m \times r}}$ and ${\bf{V}} \in {\mathbb{R}^{m \times r}}$
contains the corresponding singular vectors.
When $R$ is $C^2$ on $(0, + \infty )$ with first- and second-order derivative be $R'$ and $R''$, respectively, denote
\[{\bf{\Sigma '}} = {\bf{diag}}(R'({\sigma _1}({\bf{X}})), \cdots ,R'({\sigma _r}({\bf{X}})))\]
\[{\bf{\Sigma ''}} = {\bf{diag}}(R''({\sigma _1}({\bf{X}})), \cdots ,R''({\sigma _r}({\bf{X}}))).\]
The differential of ${\bf{X}}$ can be computed as
\begin{equation}\label{ap-25}
{\rm{d}}{\bf{X}} = {\rm{d}}{\bf{U\Sigma }}{{\bf{V}}^T} + {\bf{U}}{\rm{d}}{\bf{\Sigma }}{{\bf{V}}^T}{\rm{ + }}{\bf{U\Sigma }}{\rm{d}}{{\bf{V}}^T}.
\end{equation}
Meanwhile, with ${{\bf{U}}^T}{\bf{U}} = {{\bf{V}}^T}{\bf{V}} = {{\bf{I}}_r}$ and
\begin{equation}\label{ap-26}
{{\bf{U}}^T}{\rm{d}}{\bf{U\Sigma }} + {\bf{\Sigma }}{\rm{d}}{{\bf{V}}^T}{\bf{V}} = {\bf{0}}
\end{equation}
it follows that
\begin{equation}\label{ap-27}
{{\bf{U}}^T}{\rm{d}}{\bf{XV}} = {{\bf{U}}^T}{\rm{d}}{\bf{U\Sigma }} + {\rm{d}}{\bf{\Sigma }} + {\bf{\Sigma }}{\rm{d}}{{\bf{V}}^T}{\bf{V}} = {\rm{d}}{\bf{\Sigma }}.
\end{equation}
Then, we have
\begin{equation}\label{ap-28}
{\rm{d}}R({{\boldsymbol{\sigma }}_r}({\bf{X}})) = {\rm{d}}({\rm{tr(}}R({\bf{\Sigma }}){\rm{))}} = {\rm{tr(}}{\bf{\Sigma '}}{\rm{d}}{\bf{\Sigma }}{\rm{)}} = {\rm{tr(}}{\bf{V\Sigma '}}{{\bf{U}}^T}{\rm{d}}{\bf{X}}{\rm{)}}.
\end{equation}
Thus, the gradient of $R({{\boldsymbol{\sigma }}_r}({\bf{X}}))$ is given by
\[{\nabla _{\bf{X}}}R({{\boldsymbol{\sigma }}_r}({\bf{X}})) = {\bf{U\Sigma '}}{{\bf{V}}^T}.\]

\subsection{Derivation of (12)}

Using (\ref{ap-25})--(\ref{ap-28})),
the differential of the objective $T$ with respect to
${\bf{X}}$ can be expressed as
\begin{equation}\label{ap-29}
\begin{split}
&{\rm{d}}(T({\bf{X}}))\\
& = {\rm{d}}({\rm{tr(}}R({\bf{\Sigma }}){\rm{))}} + \frac{\eta }{2}{\rm{d}}\left( {{\rm{tr(}}{{\bf{\Sigma }}^2}{\rm{)}} - {\rm{2tr}}({{\bf{T}}^T}{\bf{U\Sigma }}{{\bf{V}}^T})} \right)\\
& = {\rm{tr(}}{\bf{V\Sigma '}}{{\bf{U}}^T}{\rm{d}}{\bf{X}}{\rm{)}} + \eta \left( {{\rm{tr(}}{\bf{\Sigma }}{\rm{d}}{\bf{\Sigma }}{\rm{)}} - {\rm{tr}}({{\bf{T}}^T}{\bf{U}}{\rm{d}}{\bf{\Sigma }}{{\bf{V}}^T})} \right)\\
& = {\rm{tr(}}{\bf{V\Sigma '}}{{\bf{U}}^T}{\rm{d}}{\bf{X}}{\rm{)}} + \eta {\rm{tr(}}{\bf{V\Sigma }}{{\bf{U}}^T}{\rm{d}}{\bf{X}}{\rm{)}} - \\ &~~~~~~~~~~~~~~~~~~~~~~~~~~~~~~\eta{\rm{tr}}({\bf{V}}{{\bf{V}}^T}{{\bf{T}}^T}{\bf{U}}{{\bf{U}}^T}{\rm{d}}{\bf{X}}) .
\end{split}
\end{equation}
Thus, we have
\[{\nabla _{\bf{X}}}T({\bf{X}}) = {{\bf{U}}^T}{\bf{\Sigma 'V}} + \eta \left( {{\bf{X}} + {\bf{U}}{{\bf{U}}^T}{\bf{TV}}{{\bf{V}}^T}} \right)\]
which results in (\ref{II-12}).

\subsection{Hessian of $R({{\boldsymbol{\sigma }}_r}({\bf{X}}))$}
Follows from (\ref{ap-28}), using (\ref{ap-25}) we have
\begin{equation}\label{ap-30}
\begin{split}
&{{\rm{d}}^2}R({{\boldsymbol{\sigma }}_r}({\bf{X}})) \\
&= {\rm{tr}}\left( {\left[ {{\rm{d}}{\bf{V\Sigma '}}{{\bf{U}}^T} + {\bf{V\Sigma ''}}{\rm{d}}{\bf{\Sigma }}{{\bf{U}}^T} + {\bf{V\Sigma '}}{\rm{d}}{{\bf{U}}^T}} \right]{\rm{d}}{\bf{X}}} \right)\\
& = {\rm{tr}}\left( {{\bf{V\Sigma ''}}{\rm{d}}{\bf{\Sigma }}{{\bf{U}}^T}{\rm{d}}{\bf{X}}} \right) + {\rm{tr}}\left( {{\rm{d}}{{\bf{X}}^T}({\rm{d}}{\bf{U\Sigma '}}{{\bf{V}}^T}{\rm{ + }}{\bf{U\Sigma '}}{\rm{d}}{{\bf{V}}^T})} \right).
\end{split}
\end{equation}
Next, we show that
\begin{equation}\label{ap-a-33}
{\rm{tr}}\left( {{\rm{d}}{{\bf{X}}^T}({\rm{d}}{\bf{U\Sigma '}}{{\bf{V}}^T}{\rm{ + }}{\bf{U\Sigma '}}{\rm{d}}{{\bf{V}}^T})} \right) = {\bf{0}}.
\end{equation}
There exists a full SVD ${\bf{X}}{\rm{ = }}\bar{\bf{ U}}\bar {\bf{\Sigma }}{\bar{\bf{ V}}^T}$, with $\bar{\bf{ U}} \in {\mathbb{R}^{m \times m}}$, $\bar{\bf{ V}} \in {\mathbb{R}^{n \times n}}$ and $\bar{\bf{ \Sigma }} \in {\mathbb{R}^{m \times n}}$, such that
\[{\bf{U}} = \bar{\bf{ U}}(:,1:r),~~{\bf{V}} = \bar{\bf{ V}}(:,1:r),~~\bar{\bf{ \Sigma }} = \left[ {\begin{array}{*{20}{c}}
\!{\bf{\Sigma }}&{\bf{0}}\!\\
\!{\bf{0}}&{\bf{0}}\!
\end{array}} \right].\]
Then, denote
\[\bar{\bf{ \Sigma}}' = \left[ {\begin{array}{*{20}{c}}
\!{{\bf{\Sigma '}}}&{\bf{0}}\!\\
\!{\bf{0}}&{\bf{0}}\!
\end{array}} \right]\]
and use ${\bar{\bf{ V}}^T}\bar{\bf{ V}} = \bar{\bf{ V}}{\bar{\bf{ V}}^T} = {{\bf{I}}_n}$,
${\bar{\bf{ U}}^T}\bar{\bf{ U}}{\rm{ = }}\bar{\bf{ U}}{\bar{\bf{ U}}^T}{\rm{ = }}{{\bf{I}}_m}$,
${\bar{\bf{ U}}^T}{\rm{d}}\bar{\bf{ U}}\bar {\bf{\Sigma}}'{\rm{ + }}\bar{\bf{ \Sigma}}'{\rm{d}}{\bar{\bf{ V}}^T}\bar{\bf{ V}} = {\bf{0}}$,
(\ref{ap-a-33}) can be justified as
\begin{align} \notag
&{\rm{tr}}\left( {{\rm{d}}{{\bf{X}}^T}\left( {{\rm{d}}{\bf{U\Sigma '}}{{\bf{V}}^T}{\rm{ + }}{\bf{U\Sigma '}}{\rm{d}}{{\bf{V}}^T}} \right)} \right)\\\notag
& = {\rm{tr}}\left( {{\rm{d}}{{\bf{X}}^T}\left( {{\rm{d}}\bar{\bf{ U}}\bar {\bf{\Sigma}}'{{\bar{\bf{ V}}}^T}{\rm{ + }}\bar{\bf{ U}}\bar {\bf{\Sigma}}'{\rm{d}}{{\bar{\bf{ V}}}^T}} \right)} \right)\\\notag
& = {\rm{tr}}\left( {{\rm{d}}{{\bf{X}}^T}\bar{\bf{ U}}{{\bar{\bf{ U}}}^T}\left( {{\rm{d}}\bar{\bf{ U}}\bar {\bf{\Sigma}}'{{{\bf{\bar V}}}^T}{\rm{ + }}\bar{\bf{ U}}\bar{\bf{ \Sigma}}'{\rm{d}}{{\bar{\bf{ V}}}^T}} \right)\bar{\bf{ V}}{{\bar{\bf{ V}}}^T}} \right)\\\notag
& = {\rm{tr}}\left( {{\rm{d}}{{\bf{X}}^T}\bar{\bf{ U}}\left( {{{\bar{\bf{ U}}}^T}{\rm{d}}\bar{\bf{ U}}\bar {\bf{\Sigma}}'{\rm{ + }}\bar{\bf{ \Sigma '}}{\rm{d}}{{\bar{\bf{ V}}}^T}\bar{\bf{ V}}} \right){{\bar{\bf{ V}}}^T}} \right)\\\notag
& = {\bf{0}}.
\end{align}
Substituting (\ref{ap-a-33}) into (\ref{ap-30}), and using (\ref{ap-27}) and
${\rm{tr}}({\bf{ABCD}}) = {\rm{ve}}{{\rm{c}}^T}({{\bf{B}}^T})({{\bf{A}}^T} \otimes {\bf{C}}){\rm{vec}}({\bf{D}})$ yield
\begin{equation}\label{ap-31}
\begin{split}
&{{\rm{d}}^2}R({{\boldsymbol{\sigma }}_r}({\bf{X}})) = {\rm{tr}}\left( {{\bf{V\Sigma ''}}{{\bf{U}}^T}{\rm{d}}{\bf{XV}}{{\bf{U}}^T}{\rm{d}}{\bf{X}}} \right)\\
& = {[{\rm{d(vec(}}{\bf{X}}{\rm{))]}}^T}{{\bf{K}}_{nm}}[({\bf{U\Sigma ''}}{{\bf{V}}^T}) \otimes ({\bf{V}}{{\bf{U}}^T})]{\rm{d(vec(}}{\bf{X}}{\rm{))}}
\end{split}
\end{equation}
where ${{\bf{K}}_{nm}}$ is a commutation matrix defined as
${\rm{vec}}({\bf{A}}) = {{\bf{K}}_{nm}}{\rm{vec}}({{\bf{A}}^T})$ for ${\bf{A}} \in {\mathbb{R}^{m \times n}}$.
Then, follows from (\ref{ap-31}) and the relation between Hessian matrix and second-order differential \cite{51},
Lemma 2 is derived.

\section{Proof of Lemma 3}

First, using $({\bf{AB}}) \otimes ({\bf{CD}}) = ({\bf{A}} \otimes {\bf{C}})({\bf{B}} \otimes {\bf{D}})$,
we have
\begin{align}\notag
&({\bf{U\Sigma ''}}{{\bf{V}}^T}) \otimes ({\bf{V}}{{\bf{U}}^T}) + ({\bf{U}}{{\bf{V}}^T}) \otimes ({\bf{V\Sigma ''}}{{\bf{U}}^T})\\\notag
& = ({\bf{U}} \otimes {\bf{V}})({\bf{\Sigma ''}} \otimes {{\bf{I}}_r}){({\bf{V}} \otimes {\bf{U}})^T}\\\notag
& ~~+ ({\bf{U}} \otimes {\bf{V}})({{\bf{I}}_r} \otimes {\bf{\Sigma ''}}){({\bf{V}} \otimes {\bf{U}})^T}\\\notag
& = ({\bf{U}} \otimes {\bf{V}})({\bf{\Sigma ''}} \otimes {{\bf{I}}_r} + {{\bf{I}}_r} \otimes {\bf{\Sigma ''}}){({\bf{V}} \otimes {\bf{U}})^T}.
\end{align}
Then, with the properties of commutation matrix,
\[{{\bf{K}}_{nm}}({\bf{U}} \otimes {\bf{V}}){{\bf{K}}_{rr}} = {\bf{V}} \otimes {\bf{U}}\]
and ${{\bf{K}}_{rr}}{\bf{K}}_{rr}^{ - 1} = {{\bf{K}}_{rr}}{{\bf{K}}_{rr}} = {{\bf{I}}_{{r^2}}}$,
it follows that
\begin{align}\notag
&\nabla _{\bf{X}}^2R({{\boldsymbol{\sigma }}_r}({\bf{X}}))\\\notag
& = \frac{1}{2}{{\bf{K}}_{nm}}\big[ {({\bf{U}} \otimes {\bf{V}})({\bf{\Sigma ''}} \otimes {{\bf{I}}_r} + {{\bf{I}}_r} \otimes {\bf{\Sigma ''}}){{({\bf{V}} \otimes {\bf{U}})}^T}} \big]\\\notag
& = \frac{1}{2}({\bf{V}} \otimes {\bf{U}})\left[ {{{\bf{K}}_{rr}}({\bf{\Sigma ''}} \otimes {{\bf{I}}_r} + {{\bf{I}}_r} \otimes {\bf{\Sigma ''}})} \right]{({\bf{V}} \otimes {\bf{U}})^T}.
\end{align}
Since ${{\bf{V}}^T}{\bf{V}} = {{\bf{U}}^T}{\bf{U}} = {{\bf{I}}_r}$,
it is easy to see that
\[{({\bf{V}} \otimes {\bf{U}})^T}({\bf{V}} \otimes {\bf{U}}) = ({{\bf{V}}^T}{\bf{V}} \otimes {{\bf{U}}^T}{\bf{U}}) = {{\bf{I}}_{{r^2}}}\]
which implies the columns of the matrix $({\bf{V}} \otimes {\bf{U}})$ are orthogonal.
Meanwhile, the commutation matrix ${{\bf{K}}_{rr}}$ is orthogonal and in fact
${{\bf{K}}_{rr}}({\bf{\Sigma ''}} \otimes {{\bf{I}}_r} + {{\bf{I}}_r} \otimes {\bf{\Sigma ''}})$
is a rearrange of the diagonal elements of the diagonal matrix
$({\bf{\Sigma ''}} \otimes {{\bf{I}}_r} + {{\bf{I}}_r} \otimes {\bf{\Sigma ''}})$.
Thus, when $R'' \neq 0$ on $(0,\infty )$,
it follows from ${{\boldsymbol{\sigma }}_r}({\bf{X}})>0$ that
\[{\rm{rank}}\left( {\nabla _{\bf{X}}^2R({{\boldsymbol{\sigma }}_r}({\bf{X}}))} \right) = {r^2}\]
and the $r^2$ nonzero eigenvalues of $\nabla _{\bf{X}}^2R({{\boldsymbol{\sigma }}_r}({\bf{X}}))$
are given by 
\[{\boldsymbol{\lambda }}\left( {\nabla _{\bf{X}}^2R({{\boldsymbol{\sigma }}_r}({\bf{X}}))} \right)
= {\boldsymbol{\lambda }}({\bf{\Sigma ''}} \otimes {{\bf{I}}_r} + {{\bf{I}}_r} \otimes {\bf{\Sigma ''}}).\]
Moreover, under the assumption that $R''$ is a nondecreasing function on $(0,\infty )$,
and with ${\sigma _r}({\bf{X}}) = \min ({{\boldsymbol{\sigma }}_r}({\bf{X}}))$,
we have
\[{\lambda _{\min }}\left( {\nabla _{\bf{X}}^2R({{\boldsymbol{\sigma }}_r}({\bf{X}}))} \right) = R''({\sigma _r}({\bf{X}}))\]
\[{\rm{and}}~~~~~~~~~~{\lambda _{\max }}\left( {\nabla _{\bf{X}}^2R({{\boldsymbol{\sigma }}_r}({\bf{X}}))} \right) = 0~~~~~~~~~~~~~~~~~~~~~~~~~~~~\]
which concludes the proof.

\section{Proof of Lemma 4}

Let ${\beta _L}$ be the larger output of the singular value thresholding function
${P_{ R,L/\lambda }}$ (corresponding to ${\bar P_{\bar R,L/\lambda }}$ in (\ref{II-16}))
at its discontinuous point. That is, ${\beta _L}$ is the jumping size
at the discontinuous point of ${P_{ R,L/\lambda }}$.
Then, for any ${{\bf{X}}^k}$ generated by the PGD algorithm,
it follows from the discontinuous thresholding property that,
for $ 1 \le i \le \min (m,n),~\forall k > 0$,
\begin{equation}\label{ap-32}
\begin{split}
{\sigma _i}({{\bf{X}}^k}) \ge {\beta _L},&~~~{\rm{if}}~~~{\sigma _i}({{\bf{X}}^k}) \ne 0.
\end{split}
\end{equation}

By Property 2(ii), there exists a sufficiently large positive
integer ${k_0}$ such that when $k > {k_0}$ it holds
\[{\left\| {{{\bf{X}}^{k + 1}} - {{\bf{X}}^k}} \right\|_F} < {\beta _L}\]
which together with Lemma 1 implies
\begin{equation}\label{ap-33}
{\left\| {{\boldsymbol{\sigma }}({{\bf{X}}^{k + 1}}) - {\boldsymbol{\sigma }}({{\bf{X}}^k})} \right\|_2} < {\beta _L}.
\end{equation}
Denote ${r^k} = {\rm{rank}}({{\bf{X}}^k})$, it follows from (\ref{ap-32}) that
\[{\left\| {{\boldsymbol{\sigma }}({{\bf{X}}^{k + 1}}) - {\boldsymbol{\sigma }}({{\bf{X}}^k})} \right\|_2} \ge {\beta _L},~~{\rm{if}}~~{r^{k + 1}} \ne {r^k}\]
which contradicts to (\ref{ap-33}) when $k > {k_0}$.
Thus, ${r^{k + 1}} = {r^k}$ when $k > {k_0}$.
It means that the rank of ${{\bf{X}}^k}$ converges
\begin{equation}\label{ap-34}
{r^{k + 1}} = {r^k} = r,~~\forall k > {k_0}.
\end{equation}

For any cluster point ${{\bf{X}}^*}$,
there exists a subsequence $\{ {{\bf{X}}^{{k_j}}}\} $ converging to ${{\bf{X}}^*}$,
i.e., ${{\bf{X}}^{{k_j}}} \to {{\bf{X}}^*}$ as $j \to \infty $.
Thus, there exists a sufficiently large positive integer ${j_0}$ such that
${k_{{j_0}}} > {k_0}$ and
\[{\left\| {{\boldsymbol{\sigma }}({{\bf{X}}^{{k_j}}}) - {\boldsymbol{\sigma }}({{\bf{X}}^*})} \right\|_2} < {\beta _L}\]
when $j > {j_0}$. Similar to the above analysis, we have
\[{r^{{k_j}}} = {\rm{rank(}}{{\bf{X}}^*}{\rm{)}},~~~~\forall j > {j_0}.\]
From (\ref{ap-34}), ${r^{{k_j}}} = r$, thus ${\rm{rank(}}{{\bf{X}}^*}{\rm{)}} = r$
for any cluster point ${{\bf{X}}^*}$. Consequently, taking ${k^*} > {k_{{j_0}}}$,
Lemma 4 is proved based on the above analysis.

\section{Proof of Theorem 1}

The condition in Theorem 1 implies that
\begin{equation}\label{ap-36}
\nabla _{\bf{X}}^2\bar F({{\bf{X}}^*}) = \lambda \nabla _{\bf{X}}^2R({{\boldsymbol{\sigma }}_r}({{\bf{X}}^*})) + {\bf{diag}}({\rm{vec}}({{\bf{{\bf P}}}_\Omega }))\succeq{\bf{0}}.
\end{equation}
Consider a sufficiently small matrix ${\bf{E}}$ with
${\left\| {\bf{E}} \right\|_F} < {\beta _L}$,
${\beta _L}$ is the is the ``jumping'' size of the
singular value thresholding function ${P_{ R,L/\lambda }}$
(corresponding to ${\bar P_{\bar R,L/\lambda }}$ in (\ref{II-16}))
at the its discontinuous point. Under Assumption 2,
we have $\min \left( {{{\boldsymbol{\sigma }}_r}({{\bf{X}}^*})} \right) \ge {\beta _L}$,
thus, ${\rm{rank}}({{\bf{X}}^*} + {\bf{E}}) \ge r$ for such a small ${\bf{E}}$.
This can be justified as follows. With ${\left\| {\bf{E}} \right\|_F} < {\beta _L}$, by Lemma 1
\begin{equation}\label{ap-3-38}
{\left\| {{{\boldsymbol{\sigma }}_r}({{\bf{X}}^*}) - {{\boldsymbol{\sigma }}_r}({{\bf{X}}^*} + {\bf{E}})} \right\|_2} <{\beta _L}.
\end{equation}
Since $\min \left( {{{\boldsymbol{\sigma }}_r}({{\bf{X}}^*})} \right) \ge {\beta _L}$,
it follows that
\[{\left\| {{{\boldsymbol{\sigma }}_r}({{\bf{X}}^*}) - {{\boldsymbol{\sigma }}_r}({{\bf{X}}^*} + {\bf{E}})} \right\|_2} \ge {\beta _L}
~~{\rm{if}}~~{\rm{rank}}({{\bf{X}}^*} + {\bf{E}}) < r\]
which contradict to (\ref{ap-3-38}).

Let $\tilde{\bf{ U}}{\bf{diag}}\left( {{\boldsymbol{\sigma }}({{\bf{X}}^*} + {\bf{E}})} \right){\tilde{\bf{ V}}^T}$
be any full SVD of $({{\bf{X}}^*} + {\bf{E}})$ and denote
\[{\bf{X}}_e^* = \tilde{\bf{ U}}{\bf{diag}}\left( {[{\boldsymbol{\sigma }}_r^T({{\bf{X}}^*} + {\bf{E}}),{\bf{0}}]} \right){\tilde{\bf{ V}}^T}\]
\[{\bf{X}}_{e \bot }^* = \tilde{\bf{ U}}{\bf{diag}}\left( {[{\bf{0}},{\boldsymbol{\sigma }}_{r \bot }^T({{\bf{X}}^*} + {\bf{E}})]} \right){\tilde{\bf{ V}}^T}.\]
From the property of stationary point, ${{\bf{X}}^*}$ satisfies
\begin{equation}\label{ap-37}
{\nabla _{\bf{X}}}G({{\bf{X}}^*}) + \lambda {\nabla _{\bf{X}}}R({{\boldsymbol{\sigma }}_r}({{\bf{X}}^*})) = {\bf{0}}.
\end{equation}
Then, it follows from (\ref{ap-36}) and (\ref{ap-37}) that
for sufficiently small matrix ${\bf{E}}$,
\begin{equation}\label{ap-38}
\begin{split}
&G({\bf{X}}_e^*) + \lambda R({{\boldsymbol{\sigma }}_r}({{\bf{X}}^*} + {\bf{E}}))\\
& \ge G({{\bf{X}}^*}) + \lambda R({{\boldsymbol{\sigma }}_r}({{\bf{X}}^*})) = F({{\bf{X}}^*}).
\end{split}
\end{equation}
Denote
\[{\bf{y}} = {\rm{diag}}\left( {{{\tilde{\bf{ V}}}^T}{{[{\nabla _{{\bf{X}}_{e \bot }^*}}f({\bf{X}}_e^*)]}^T}\tilde{\bf{ U}}} \right).\]
For sufficiently small ${\bf{E}}$, by Lemma 1 and
${\rm{rank}}({{\bf{X}}^*}) = r$, ${\sigma _i}({{\bf{X}}^*} + {\bf{E}})$
is also sufficiently small for $r + 1 \le i \le \min (m,n)$,
then under Assumption 1 it holds that for $r + 1 \le i \le \min (m,n)$,
\[R({\sigma _i}({{\bf{X}}^*} + {\bf{E}})) \ge \frac{{{{\left\| {\bf{y}} \right\|}_\infty }}}{\lambda }{\sigma _i}({{\bf{X}}^*} + {\bf{E}})\]
where the equality holds if and only if ${\sigma _i}({{\bf{X}}^*} + {\bf{E}}) = 0$.
Thus, for a sufficiently small ${\bf{E}}$
(hence ${\sigma _i}({{\bf{X}}^*} + {\bf{E}})$ is sufficient small for
$r + 1 \le i \le \min (m,n)$), using
${{\bf{X}}^*} + {\bf{E}} = {\bf{X}}_e^* + {\bf{X}}_{e \bot }^*$ and
${\bf{X}}_{e \bot }^*$ be also sufficient small, it holds that
\begin{equation}\label{ap-39}
\begin{split}
&G({{\bf{X}}^*} + {\bf{E}}) - G({\bf{X}}_e^*) + \lambda R({{\boldsymbol{\sigma }}_{r \bot }}({{\bf{X}}^*} + {\bf{E}}))\\
& = \left\langle {{\nabla _{{\bf{X}}_{e \bot }^*}}G({\bf{X}}_e^*),{\bf{X}}_{e \bot }^*} \right\rangle  + \lambda R({{\boldsymbol{\sigma }}_{r \bot }}({{\bf{X}}^*} + {\bf{E}})) + o\left( {{{\left\| {{\bf{X}}_{e \bot }^*} \right\|}_F}} \right)\\
& = {\rm{tr}}\left( {{{[{\nabla _{{\bf{X}}_{e \bot }^*}}G({\bf{X}}_e^*)]}^T}\tilde{\bf{ U}}{\bf{diag}}\left( {[{\bf{0}},{\boldsymbol{\sigma }}_{r \bot }^T({{\bf{X}}^*} + {\bf{E}})]} \right){{\tilde{\bf{ V}}}^T}} \right)\\
&~~~~~~~~~~ + \lambda R({{\boldsymbol{\sigma }}_{r \bot }}({{\bf{X}}^*} + {\bf{E}})) + o\left( {{{\left\| {{\boldsymbol{\sigma }}_{r \bot }^T({{\bf{X}}^*} + {\bf{E}})} \right\|}_2}} \right)\\
& = \sum\limits_{i = r + 1}^{\min (m,n)} {[{\bf{y}}(i){\sigma _i}({{\bf{X}}^*} + {\bf{E}}) + \lambda R({\sigma _i}({{\bf{X}}^*} + {\bf{E}}))]} \\
&~~~~~~~~~~~~~~~~~ + o\left( {{{\left\| {{\boldsymbol{\sigma }}_{r \bot }^T({{\bf{X}}^*} + {\bf{E}})} \right\|}_2}} \right)\\
& \ge 0.
\end{split}
\end{equation}
Then, summing up the two inequalities (\ref{ap-38}) and (\ref{ap-39}), we have
\[F({{\bf{X}}^*} + {\bf{E}}) - F({{\bf{X}}^*}) \ge 0\]
for sufficiently small ${\bf{E}}$, which implies that ${{\bf{X}}^*}$ is a local minimizer of $F$.

\section{Proof of Theorem 2}

The derivation follows similar to that in Appendix D.
Briefly, the condition in Theorem 2 implies that
\begin{equation}\label{ap-4-42}
\nabla _{{{\bf{X}}_\Omega }}^2\bar F({{\bf{X}}^*}) = \lambda \nabla _{{{\bf{X}}_\Omega }}^2R({{\boldsymbol{\sigma }}_r}({{\bf{X}}^*})) + {{\bf{I}}_{|\Omega |}} \succ {\bf{0}}.
\end{equation}
Consider a sufficiently small matrix ${\bf{E}}$ with
${\left\| {\bf{E}} \right\|_F} < {\beta _L}$ such that
${\rm{rank}}({{\bf{X}}^*} + {\mathcal{P}_\Omega }({\bf{E}})) \ge r$ under Assumption 2.
Let $\tilde{\bf{ U}}{\bf{diag}}\left( {{\boldsymbol{\sigma }}({{\bf{X}}^*} + {\mathcal{P}_\Omega }({\bf{E}}))} \right){\tilde{\bf{ V}}^T}$
be any full SVD of $({{\bf{X}}^*} + {\mathcal{P}_\Omega }({\bf{E}}))$ and denote
\[{\bf{X}}_e^* = \tilde{\bf{ U}}{\bf{diag}}\left( {[{\boldsymbol{\sigma }}_r^T({{\bf{X}}^*} + {\mathcal{P}_\Omega }({\bf{E}})),{\bf{0}}]} \right){\tilde{\bf{ V}}^T}\]
\[{\bf{X}}_{e \bot }^* = \tilde{\bf{ U}}{\bf{diag}}\left( {[{\bf{0}},{\boldsymbol{\sigma }}_{r \bot }^T({{\bf{X}}^*} + {\mathcal{P}_\Omega }({\bf{E}}))]} \right){\tilde{\bf{ V}}^T}.\]
From the property of stationary point, ${{\bf{X}}^*}$ satisfies
\begin{equation}\label{ap-4-43}
{\nabla _{{{\bf{X}}_\Omega }}}G({{\bf{X}}^*}) + \lambda {\nabla _{{{\bf{X}}_\Omega }}}R({{\boldsymbol{\sigma }}_r}({{\bf{X}}^*})) = {\bf{0}}.
\end{equation}
Then, it follows from (\ref{ap-4-42}) and (\ref{ap-4-43}) that
for sufficiently small matrix ${\bf{E}}$,
\begin{equation}\label{ap-4-44}
\begin{split}
&G({\bf{X}}_e^*) + \lambda R({{\boldsymbol{\sigma }}_r}({{\bf{X}}^*} + {\mathcal{P}_\Omega }({\bf{E}})))\\
& > G({{\bf{X}}^*}) + \lambda R({{\boldsymbol{\sigma }}_r}({{\bf{X}}^*})) = F({{\bf{X}}^*}).
\end{split}
\end{equation}
For sufficiently small ${\bf{E}}$,
${\sigma _i}({{\bf{X}}^*} + {\bf{E}})$ is also sufficiently
small for $r + 1 \le i \le \min (m,n)$, then, similar to (\ref{ap-39}) we have
\begin{equation}\label{ap-4-45}
G({{\bf{X}}^*} + {\mathcal{P}_\Omega }({\bf{E}})) - G({\bf{X}}_e^*) + \lambda R({{\boldsymbol{\sigma }}_{r \bot }}({{\bf{X}}^*} + {\mathcal{P}_\Omega }({\bf{E}}))) \ge 0.
\end{equation}
Then, summing up (\ref{ap-4-44}) and (\ref{ap-4-45}),
it follows that for sufficiently small ${\bf{E}}$,
\[F({{\bf{X}}^*} + {\mathcal{P}_\Omega }({\bf{E}})) - F({{\bf{X}}^*}) > 0\]
which implies that ${{\bf{X}}^*}$ is a
$\Omega$-restricted strictly local minimizer of $F$ by Definition 2.

\section{Proof of Theorem 3}
From Lemma 4, for $\delta  < {\beta _L}$,
there exists a sufficiently large integer ${k^0} > {k^*}$
(${k^*}$ defined in Lemma 4) such that
${\left\| {{{\bf{X}}^k} - {{\bf{X}}^*}} \right\|_F} < \delta $ and
${\rm{rank}}({{\bf{X}}^k}) = {r}$, $\forall k > {k^0}$.
Let ${\bf{X}}$ be a rank-$r$ matrix with a truncated SVD
${\bf{X}} = {\bf{Udiag}}\left( {{{\boldsymbol{\sigma }}_r}({\bf{X}})} \right){{\bf{V}}^T}$,
by Lemma 4, when $k > {k^0}$ the PGD algorithm in fact minimizes the following objective
\[f({\bf{X}}) := \lambda R({{\boldsymbol{\sigma }}_r}({\bf{X}})) + \frac{L}{2}\left\| {{\bf{X}} - {{\bf{X}}^k} + \frac{1}{L}\nabla G({{\bf{X}}^k})} \right\|_F^2\]
for which the gradient is (a similar derivation as in Appendix A)
\begin{equation}\label{ap-40}
{\nabla _{\bf{X}}}f({\bf{X}}) = \lambda {\nabla _{\bf{X}}}R({{\boldsymbol{\sigma }}_{{r}}}({\bf{X}})) + L({\bf{X}} - {\bf{U}}{{\bf{U}}^T}{{\bf{Q}}^k}{\bf{V}}{{\bf{V}}^T})
\end{equation}
where ${{\bf{Q}}^k} = {{\bf{X}}^k} - \frac{1}{L}\nabla G({{\bf{X}}^k})$.
For $k > {k^0}$, let ${{\bf{X}}^k} = {{\bf{U}}^k}{\bf{diag}}({{\boldsymbol{\sigma }}_r}({{\bf{X}}^k})){({{\bf{V}}^k})^T}$
and ${{\bf{X}}^*} = {{\bf{U}}^*}{\bf{diag}}({{\boldsymbol{\sigma }}_r}({{\bf{X}}^*})){{\bf{V}}^*}^T$
be any truncated SVD of ${{\bf{X}}^k}$ and ${{\bf{X}}^*}$, respectively.
For notation simplification in the sequel, we denote
\[{\boldsymbol{\sigma }}_r^k = {{\boldsymbol{\sigma }}_r}({{\bf{X}}^k}),~{\boldsymbol{\sigma }}_r^* = {{\boldsymbol{\sigma }}_r}({{\bf{X}}^*}),~ {{\bf{Q}}^*} = {{\bf{X}}^*} - \frac{1}{L}\nabla G({{\bf{X}}^*}),\]
\[{{\bf{\Sigma }}^{k + 1}} = {({{\bf{U}}^{k + 1}})^T}{{\bf{Q}}^k}{{\bf{V}}^{k + 1}},~~{{\bf{\Sigma }}^*} = {{\bf{U}}^*}^T{{\bf{Q}}^*}{{\bf{V}}^*}.\]
From (\ref{ap-40}) the minimizer ${{\bf{X}}^{k + 1}}$ satisfies
${\nabla _{\bf{X}}}f({{\bf{X}}^{k + 1}}) = {\bf{0}}$, hence
\begin{equation}\label{ap-41}
{{\bf{X}}^{k + 1}} + \frac{\lambda}{L} {\nabla _{\bf{X}}}R({\boldsymbol{\sigma }}_r^{k + 1}) = {{\bf{U}}^{k + 1}}{{\bf{\Sigma }}^{k + 1}}{({{\bf{V}}^{k + 1}})^T}.
\end{equation}
Meanwhile,
\begin{equation}\label{ap-42}
{{\bf{X}}^*} + \frac{\lambda}{L} {\nabla _{\bf{X}}}R({\bf{\sigma }}_r^*) = {{\bf{U}}^*}{{\bf{\Sigma }}^*}{({{\bf{V}}^*})^T}.
\end{equation}
Then, it follows from (\ref{ap-41}) and (\ref{ap-42}) that
\begin{equation}\label{ap-43}
\begin{split}
&{{\bf{X}}^{k + 1}} - {{\bf{X}}^*} + \frac{\lambda}{L}[{\nabla _{\bf{X}}}R({\boldsymbol{\sigma }}_r^{k + 1}) - {\nabla _{\bf{X}}}R({\boldsymbol{\sigma }}_r^*)]\\
& = {{\bf{U}}^{k + 1}}{{\bf{\Sigma }}^{k + 1}}{({{\bf{V}}^{k + 1}})^T} - {{\bf{U}}^*}{{\bf{\Sigma }}^*}{{\bf{V}}^*}^T.
\end{split}
\end{equation}
By (\ref{III-25})
\begin{equation}\label{ap-44}
\begin{split}
&\Big\langle {{{\bf{X}}^{k + 1}}\! -\! {{\bf{X}}^*} \!+\! \frac{\lambda}{L}[{\nabla _{\bf{X}}}R({\boldsymbol{\sigma }}_{{r}}^{k + 1}) \!-\! {\nabla _{\bf{X}}}R({\boldsymbol{\sigma }}_{{r}}^*)],{{\bf{X}}^{k + 1}} - {{\bf{X}}^*}} \Big\rangle \\
& \ge (1 + \lambda R''(\sigma )/L - \lambda {c_R}/L)\left\| {{{\bf{X}}^{k + 1}} - {{\bf{X}}^*}} \right\|_F^2.
\end{split}
\end{equation}
From Property 1, ${{\bf{U}}^{k + 1}}$ and ${{\bf{V}}^{k + 1}}$ are
the singular vectors of ${{\bf{Q}}^k}$ corresponding to ${{\boldsymbol{\sigma }}_r}({{\bf{Q}}^k})$,
and
\[{{\bf{\Sigma }}^{k + 1}} = {({{\bf{U}}^{k + 1}})^T}{{\bf{Q}}^k}{{\bf{V}}^{k + 1}} = {\bf{diag}}({{\boldsymbol{\sigma }}_r}({{\bf{Q}}^k})).\]
Meanwhile, ${{\bf{U}}^{*}}$ and ${{\bf{V}}^{*}}$ are the singular vectors of
${{\bf{Q}}^*}$ corresponding to ${{\boldsymbol{\sigma }}_r}({{\bf{Q}}^*})$,
and
\[{{\bf{\Sigma }}^*} = {{\bf{U}}^*}^T{{\bf{Q}}^*}{{\bf{V}}^*} = {\bf{diag}}({{\boldsymbol{\sigma }}_r}({{\bf{Q}}^*})).\]
Then, it follows from (\ref{III-26}) and Assumption 3 that,
in a sufficiently small neighborhood of ${{\bf{X}}^*}$,
there exists constants ${\alpha ^k} := \alpha ({{\bf{X}}^k})$ (which is sufficiently small),
${\gamma ^k} := \gamma ({{\bf{X}}^k}) \in [0,1)$ and
${\xi ^k} := \xi ({{\bf{X}}^k}) \in (0,1)$,
satisfying ${\beta ^k}: = 1/(1 + 2{\alpha ^k}) - {\gamma ^k} > 0$, such that
\begin{equation}\label{ap-45}
\begin{split}
&\left\| {{{\bf{U}}^{k + 1}}{{\bf{\Sigma }}^{k + 1}}{{({{\bf{V}}^{k + 1}})}^T} - {{\bf{U}}^*}{{\bf{\Sigma }}^*}{{\bf{V}}^*}^T} \right\|_F^2\\
& = {\beta ^k}\left\| {{{\bf{Q}}^k} - {{\bf{Q}}^*}} \right\|_F^2\\
& \!= \!{\beta ^k}\left\| {\Big(1 \!-\! \frac{1}{L}\Big)\left[{{\mathcal{P}}_\Omega }({{\bf{X}}^k}) \!-\! {{\mathcal{P}}_\Omega }({{\bf{X}}^*})\right] \!+\! {\mathcal{P}}_\Omega ^ \bot ({{\bf{X}}^k}) \!-\! { \mathcal{P}}_\Omega ^ \bot ({{\bf{X}}^*})} \right\|_F^2\\
& = {\beta ^k}\left\| {\Big(1 - \frac{1}{L}\Big){{\mathcal{P}}_\Omega }({{\bf{X}}^k} - {{\bf{X}}^*}) + {\mathcal{P}}_\Omega ^ \bot ({{\bf{X}}^k} - {{\bf{X}}^*})} \right\|_F^2\\
& = {\beta ^k}\left[ {{{\Big(1 - \frac{1}{L}\Big)}^2}\left\| {{{\mathcal{P}}_\Omega }({{\bf{X}}^k} - {{\bf{X}}^*})} \right\|_F^2 + \left\| {{\mathcal{P}}_\Omega ^ \bot ({{\bf{X}}^k} - {{\bf{X}}^*})} \right\|_F^2} \right]\\
& = {\beta ^k}\Big(1 -\frac{2{\xi ^k}}{L} + \frac{\xi ^k}{L^2}\Big)\left\| {{{\bf{X}}^k} - {{\bf{X}}^*}} \right\|_F^2
\end{split}
\end{equation}
where $0 < 1 - \frac{2{\xi ^k}}{L} + \frac{\xi ^k}{L^2} < 1$ since $0 < {\xi ^k} < 1$ and $L > 1$.
Then, it follows that
\begin{equation}\label{ap-46}
\begin{split}
&\big\langle {{{\bf{U}}^{k + 1}}{{\bf{\Sigma }}^{k + 1}}{{({{\bf{V}}^{k + 1}})}^T} - {{\bf{U}}^*}{{\bf{\Sigma }}^*}{{\bf{V}}^*}^T,{{\bf{X}}^{k + 1}} - {{\bf{X}}^*}} \big\rangle \\
& \le {\left\| {{{\bf{U}}^{k + 1}}{{\bf{\Sigma }}^{k + 1}}{{({{\bf{V}}^{k + 1}})}^T} - {{\bf{U}}^*}{{\bf{\Sigma }}^*}{{\bf{V}}^*}^T} \right\|_F}{\left\| {{{\bf{X}}^{k + 1}} - {{\bf{X}}^*}} \right\|_F}\\
& \le \sqrt {{\beta ^k}\Big(1 -\frac{2\xi ^k}{L} + \frac{\xi ^k}{L^2}\Big)} {\left\| {{{\bf{X}}^k} - {{\bf{X}}^*}} \right\|_F}{\left\| {{{\bf{X}}^{k + 1}} - {{\bf{X}}^*}} \right\|_F}.
\end{split}
\end{equation}
Under the conditions in Theorem 2,
we have $1 + \lambda R''(\sigma )/L > 0$
since $1 + \lambda R''(\sigma ) > 0$ and $L>1$, which implies
\[1 + \lambda R''(\sigma )/L - \lambda {c_R}/L > 0\]
for sufficiently small ${c_R}$. In this case, from (\ref{ap-43}), (\ref{ap-44}) and (\ref{ap-46}),
and without loss of any generality assuming that
${\left\| {{{\bf{X}}^{k + 1}} - {{\bf{X}}^*}} \right\|_F} > 0$ (the condition before convergence), we have
\begin{equation}\label{ap-47}
{\left\| {{{\bf{X}}^{k + 1}} - {{\bf{X}}^*}} \right\|_F} \le \frac{{\sqrt {{\beta ^k}(1 - 2{\xi ^k}{\rm{/}}L + {\xi ^k}{\rm{/}}{L^2})} }}{{1 + \lambda R''(\sigma )/L - \lambda {c_R}/L}}{\left\| {{{\bf{X}}^k} - {{\bf{X}}^*}} \right\|_F}.
\end{equation}
Let
\[{\rho ^k} = \frac{{\sqrt {{\beta ^k}(1 - 2{\xi ^k}{\rm{/}}L + {\xi ^k}{\rm{/}}{L^2})} }}{{1 + \lambda R''(\sigma )/L - \lambda {c_R}/L}}.\]
Consider a sufficiently small neighborhood of ${\bf{X}}^*$ with sufficiently small $\delta $,
thus ${c_R}$ and ${\alpha ^k}$ are sufficiently small, and with $0 \le {\gamma ^k} < 1$ and $0 < {\xi ^k} < 1$,
it holds $0 < {\rho ^k} < 1$ if
\[1 + \lambda R''(\sigma )/L > \sqrt {(1 - {\gamma ^k})(1 - 2{\xi ^k}{\rm{/}}L + {\xi ^k}{\rm{/}}{L^2})}. \]
When ${\gamma ^k}$ and ${\xi ^k}$ are respectively lower bounded by some
$\gamma  \in [0,1)$ and $\xi  \in (0,1)$, $\forall k > {k^0}$,
${\rho ^k}$ is upper bounded by some $\rho  \in (0,1)$ if
\begin{equation}\label{ap-48}
1 + \lambda R''(\sigma )/L > \sqrt {(1 - \gamma )(1 - 2\xi {\rm{/}}L + \xi {\rm{/}}{L^2})} .
\end{equation}
Thus, Theorem 3 is proved.

\end{document}